\documentclass[journal]{IEEEtran}
\usepackage{amsmath,amsfonts}
\usepackage{algorithmic}
\usepackage{algorithm}
\usepackage{amsmath}
\usepackage{amssymb}
\usepackage{array}
\usepackage[caption=false,font=normalsize,labelfont=sf,textfont=sf]{subfig}
\usepackage{textcomp}
\usepackage{stfloats}
\usepackage{url}
\usepackage{verbatim}
\usepackage{graphicx}
\usepackage{cite}
\usepackage{comment}
\usepackage{booktabs}
\usepackage{multirow}
\usepackage{dsfont}
\usepackage{tikz}
\usepackage{hyperref}

\def\checkmark{\tikz\fill[scale=0.4](0,.35) -- (.25,0) -- (1,.7) -- (.25,.15) -- cycle;} 
\newcommand{\mcrot}[3]{\multicolumn{#1}{#2}{\rlap{\rotatebox{60}{#3}}}}

\usepackage{colortbl} %JED

\begin{document}

\title{3DLabelProp: Geometric-Driven Domain Generalization for LiDAR Semantic Segmentation in Autonomous Driving}

\author{Jules Sanchez$^1$, Jean-Emmanuel Deschaud*$^1$, François Goulette$^{1,2}$
\thanks{*corresponding author}
\thanks{$^1$Centre for Robotics, Mines Paris - PSL, PSL University, Paris, France. firstname.surname@minesparis.psl.eu}
\thanks{$^2$U2IS, ENSTA Paris, Institut Polytechnique de Paris, Palaiseau, France. firstname.surname@ensta-paris.fr}}

%\markboth{IEEE Transactions on Pattern Analysis and Machine Intelligence}%
%{J. Sanchez \MakeLowercase{\textit{et al.}}: 3DLabelProp: Geometric-Driven Domain Generalization for LiDAR Semantic and Moving Object Segmentation in Autonomous Driving}


\maketitle

\begin{abstract}
Domain generalization aims to find ways for deep learning models to maintain their performance despite significant domain shifts between training and inference datasets. This is particularly important for models that need to be robust or are costly to train. LiDAR perception in autonomous driving is impacted by both of these concerns, leading to the emergence of various approaches. This work addresses the challenge by proposing a geometry-based approach, leveraging the sequential structure of LiDAR sensors, which sets it apart from the learning-based methods commonly found in the literature. The proposed method, called 3DLabelProp, is applied on the task of LiDAR Semantic Segmentation (LSS). Through extensive experimentation on seven datasets, it is demonstrated to be a state-of-the-art approach, outperforming both naive and other domain generalization methods.

The code is available on GitHub at: \\ 
\url{https://github.com/JulesSanchez/3DLabelProp}
\end{abstract}

\begin{IEEEkeywords}
Semantic Scene Understanding, LiDAR Perception, Computer Vision for Transportation, Deep Learning in Robotics and Automation, 3D Computer Vision
\end{IEEEkeywords}

\begin{figure*}[!ht]
    \centering
    \includegraphics[width=0.9\linewidth,trim={0cm 3cm 0cm 0cm},clip]{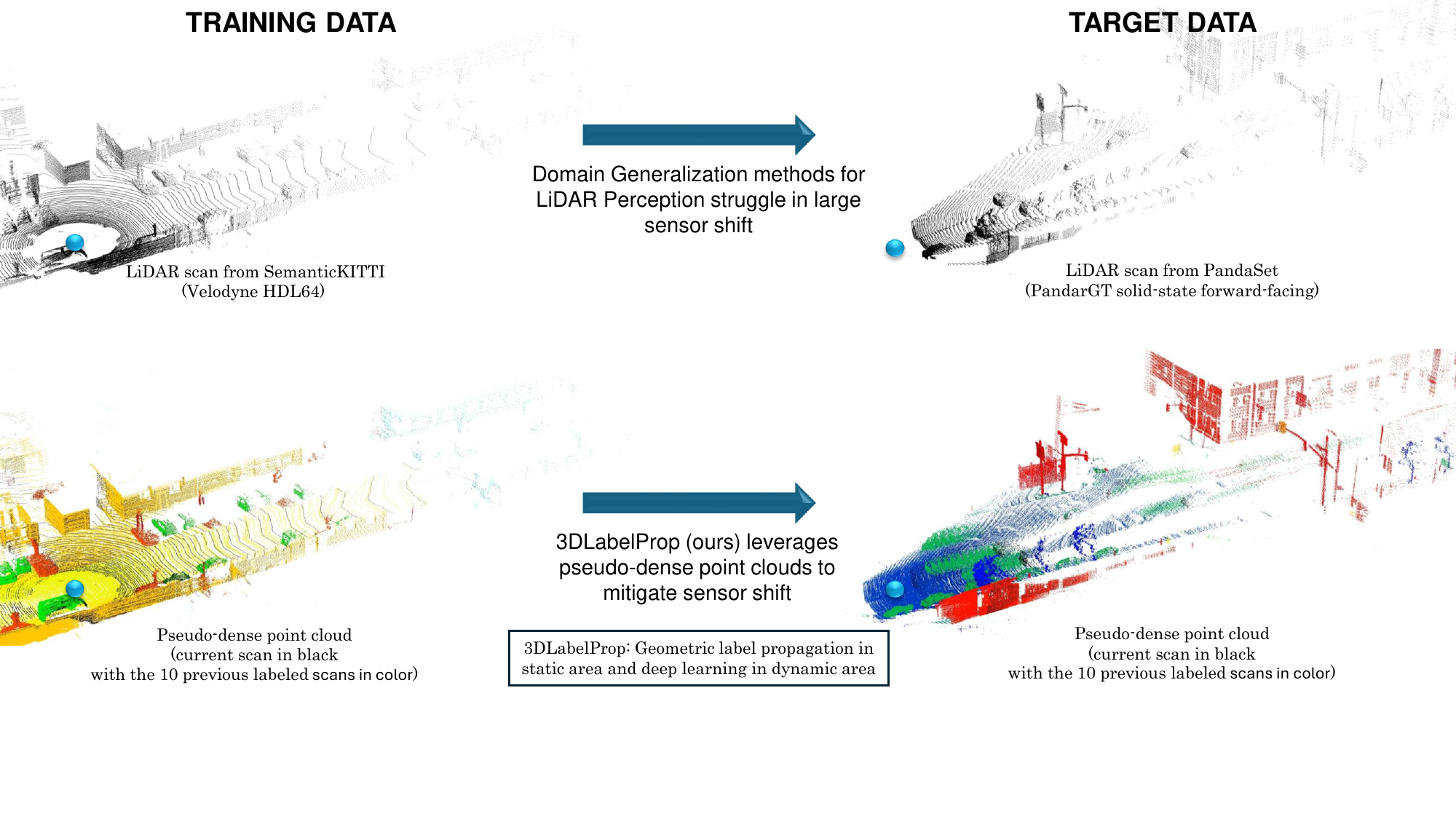}
    \caption{Illustration of our approach using pseudo-dense points for domain generalization of LiDAR semantic Segmentation in autonomous driving (the blue sphere represents the position of the ego vehicle).}
    \label{fig:teaser}
\end{figure*}

\section{Introduction}
%intro domain gen
\IEEEPARstart{D}{omain} generalization has garnered significant attention in LiDAR semantic segmentation as deep learning methods have achieved satisfactory performance on individual datasets. With the growing number of available datasets, there is an increasing need for models capable of performing cross-dataset segmentation. 

Existing methods~\cite{domgen,lidog,3DVfield,completelabel} have been predominantly learning-based, employing strategies during the training phase to enhance domain generalization performance. However, due to the challenges in evaluating cross-domain semantic segmentation, each method introduces its own set of labels, making cross-method comparisons difficult. Consequently, it becomes challenging for new research to draw insights from these approaches.

%Papier ICCV
In our previous work~\cite{3dlabelprop}, titled 'Domain Generalization of 3D Semantic Segmentation in Autonomous Driving', we made two key contributions: we introduced the first benchmark for LiDAR semantic segmentation domain generalization and proposed an approach to tackle this problem. Our new method, 3DLabelProp, is based on the idea that domain generalization can be enhanced through geometry-based strategies rather than learning-based ones, resulting in more intuitive techniques.

%Différences avec le papier
In this work, we build upon our previous research~\cite{3dlabelprop} by formalizing our experimental setup, outlining the hypotheses and decisions made. For that purpose, we introduce pseudo-dense point clouds, a foundational element of 3DLabelProp, and analyze their strengths and limitations (~\autoref{fig:teaser}). This allows us to provide a more detailed explanation and evaluation of 3DLabelProp, along with new ablation studies. We also extend the domain generalization benchmark to include more datasets, increasing from five to seven target LiDAR datasets, leading to more comprehensive conclusions. 

%Lastly, we apply 3DLabelProp to a new task—moving object segmentation—which offers valuable insights into the effectiveness of pseudo-dense methods.

%Recap contrib
Our contributions in this work are as follows:
\begin{itemize}
    \item Introduce concepts and terminology to facilitate the understanding and evaluation of domain generalization in outdoor LiDAR perception,
    \item Investigate pseudo-dense point clouds, highlighting their strengths and limitations for domain generalization in LiDAR Semantic Segmentation (LSS),
    \item Benchmark the current state-of-the-art neural models and domain generalization methods in LSS,
    \item Provide new insights into the 3DLabelProp method, offering efficient processing of pseudo-dense point clouds.
\end{itemize}

%Our contributions in this work are as follows:
%\begin{itemize}
%    \item Introduce concepts and terminology to facilitate the understanding and evaluation of domain generalization in outdoor LiDAR perception,
%    \item Investigate pseudo-dense point clouds, highlighting their strengths and limitations for domain generalization in both LiDAR Semantic Segmentation (LSS) and LiDAR Moving Object Segmentation LMOS),
%    \item Benchmark the current state-of-the-art neural models and domain generalization methods in LSS and LMOS,
%    \item Provide new insights into the 3DLabelProp method, offering faster and more efficient processing of pseudo-dense point clouds.
%\end{itemize}

\section{Related work}
\subsection{LiDAR Semantic Segmentation (LSS)}

Because point clouds lack an inherent order, traditional image-based vision techniques cannot be directly applied to LiDAR data.

The first type of deep learning method relies on permutation-invariant operations to process point clouds without requiring pre-processing. MLP-based methods~\cite{pointnet, point++, randlanet} apply a shared MLP at the point level. Other approaches redefine order-invariant convolutions~\cite{xu2018spidercnn, kpconv}. These methods often involve extensive neighborhood computations, making them time-consuming and better suited for offline point cloud semantic segmentation. Among them, KPConv~\cite{kpconv} stands out as one of the top-performing techniques.

Another type of method restructures point clouds to obtain an ordered representation. Some approaches project LiDAR point clouds into 2D, such as range-based methods~\cite{rangenet++, squeezesegv3, cenet} and bird's-eye-view methods~\cite{polarnet}, which are typically extremely fast. Others represent point clouds in a 3D regular grid and apply 3D convolutions, particularly using sparse convolutions like SRU-Net~\cite{mink}, which reduce the memory consumption of voxel-based methods. Sparse convolutions have been extended to cylindrical voxels in Cylinder3D~\cite{cylinder3d} and mixed representations, such as point-voxel methods as SPVCNN~\cite{spvnas}. Although slower than 2D methods, voxel-based approaches offer higher accuracy at reasonable speeds, making them the preferred choice for LiDAR semantic segmentation.

\subsection{Sequence-based semantic segmentation}
\label{sec:sequence-based}
Previously mentioned methods processed LiDAR scans individually. However, for autonomous driving, the data is more like a point cloud stream, similar to a video. Instead of treating each scan as separate data, this input can be viewed as a sequence of point clouds. Approaches that utilize this representation are referred to as sequence-based or 4D-based methods.

Among these methods, ASAP-Net~\cite{cao2020asap} separates spatial and temporal interactions using a temporal attention mechanism between consecutive frames. SpSequenceNet~\cite{shi2020spsequencenet} combines feature maps from previous point clouds with the current one. In~\cite{duerr2020lidar}, an RNN is employed to retain past information.

While earlier methods integrated temporal information at the feature level, others have incorporated it at the geometric level. MeteorNet~\cite{MeteorNet} and PSTNet~\cite{fan2021pstnet} redefine convolutions to account for past points in their computations. Helix4D~\cite{helix4D} represents a point cloud sequence as a hyper-cylinder, processing points within this newly defined space. \cite{aygun20214d} and \cite{kreuzberg2022stop} utilize SLAM to align each point cloud within the same reference frame, treating all points as part of a unified 3D space, with temporality represented as an input feature.

Similarly to this latter approach, 3DLabelProp~\cite{3dlabelprop} processes the registered sequence as a single 3D point cloud and applies existing methods from dense point cloud literature to perform segmentation.

\subsection{LiDAR domain generalization}
\label{sec:DG_litt}
Domain generalization has been extensively studied in machine learning and applied across various deep learning domains, including NLP and 2D computer vision. Since these areas are beyond the scope of this work, we refer the reader to~\cite{generalizationsurvey,generalizationsurvey2} for a comprehensive review. Here, we will adopt the typology they present to distinguish between different approaches to domain generalization.

Approaches can be distinguished based on two key factors: the data available for training and the strategies employed to enhance generalization performance.

In terms of data availability, methods are categorized as either single-source or multi-source. As for generalization strategies, the main approaches include meta-learning~\cite{maml}, multi-task learning~\cite{jigsaw}, data augmentation~\cite{adversarial}, neural architecture design~\cite{ibnnet}, and domain alignment~\cite{mixture}. All of these strategies have been extensively explored in the context of 2D tasks.

In the context of LiDAR scene understanding for autonomous driving, only a few works have focused on domain generalization. Specifically, in semantic segmentation, notable methods include Complete\&Label~\cite{completelabel}, 3D-VField~\cite{3DVfield,vfieldlss}, DGLSS~\cite{domgen}, LIDOG~\cite{lidog}, and COLA~\cite{tro:cola}. COLA~\cite{tro:cola} is a multi-source approach, which creates a larger dataset by relabelling and concatenating various datasets. They highlight that this larger diversity in the training set improves generalization performance. MDT3D~\cite{mdt3d} also utilizes multiple datasets simultaneously but is specifically designed for object detection. Additionally, \cite{robustness} explores the robustness of methods under degraded conditions, though we will not delve into the details of this particular work due to its limited scope on synthetic datasets built from SemanticKITTI~\cite{semantickitti}. 

Among these approaches, the three methods most closely related to ours are Complete\&Label~\cite{completelabel}, DGLSS~\cite{domgen}, and LiDOG~\cite{lidog}. We will provide a more detailed description of these methods and compare them to 3DLabelProp:
\begin{itemize}
    \item Complete\&Label (C\&L)~\cite{completelabel} focuses on domain alignment by learning a completion module, enabling the processing of scans within a canonical domain, specifically the completed domain.
    \item DGLSS~\cite{domgen} combines domain alignment and data augmentation strategies. It introduces LiDAR line dropout during training and enforces domain alignment with the original scan. Additionally, it implements IBN-Net~\cite{ibnnet} and MLDG~\cite{mldg} for 3D semantic segmentation, highlighting their inefficiency and underscoring the need for methods specifically designed for 3D.
    \item LiDOG~\cite{lidog} is a multi-task method that tries to leverage similarity between scans acquired by different sensors in a low-resolution bird's-eye-view map to help generalize a voxel-based semantic segmentation model.
\end{itemize}

%\subsection{LiDAR Moving Object Segmentation (LMOS)}

%In addition to semantic segmentation, we are also interested in the related task of LiDAR Moving Object Segmentation (LMOS). Since LMOS focuses on identifying moving objects, it analyzes the data stream rather than individual scans, similar to 4D semantic segmentation.

%LMNet~\cite{LMNET} introduced the first deep learning-based method for LMOS, along with an online benchmark using the SemanticKITTI dataset~\cite{semantickitti}. It utilizes residual images of range data, achieving very high computational speed.

%Building on this work, MotionSeg3D~\cite{sun2022efficient} and RVMOS~\cite{RVMOS} improved performance by refining the use of residual images for feature computation.

%3DSeq-MOS~\cite{3DSeqMOS} and 4DMOS~\cite{4DMOS} directly leverage 4D point clouds, while MapMOS~\cite{mersch2023building} constructs a geometric map, similar to those used in SLAM methods, alongside a 4D SRU-Net~\cite{mink}. Additionally, MapMOS addresses domain generalization performance and introduces the first benchmark for LMOS domain generalization using the nuScenes~\cite{nuscenes} and Apollo~\cite{automos} datasets.

%In this work, we propose a variation of 3DLabelProp~\cite{3dlabelprop} to tackle LMOS thanks to the native 4D approach of our method.

%\begin{figure*}[h!]
%    \centering
%    \includegraphics[width=0.7\linewidth]{images/DAvsDG.png}
%    \caption{Illustration of the difference between domain generalization and domain adaptation.}
%    \label{fig:DGvsDA}
%\end{figure*}

\section{Domain generalization for 3D}
\subsection{Introduction}

Domain generalization refers to a model's ability to perform well not only on domains encountered during training but also on new, unseen domains during inference. In 3D, domain generalization has received relatively little attention, particularly when compared to 3D domain adaptation. The key difference between the two fields is that domain adaptation assumes access to examples from the target distribution for fine-tuning the model.
%A visual example can be found in \autoref{fig:DGvsDA}.

In the field of 2D images, domain differences can arise from from variations in color, lighting conditions, seasonal changes, differences in viewpoint, or the presence of different types of objects in the scene. These variations are collectively referred to as~\textit{domain shifts}.

In the case of semantic segmentation of LiDAR data, domain shifts differ somewhat from those in 2D images from cameras. LiDAR is inherently invariant to changes in illumination and color. LiDAR domain shifts can generally be grouped into three main categories:
\begin{itemize}
\item Appearance shift,
\item Scene shift,
\item Sensor shift.
\end{itemize}

\paragraph{Appearance shift} it encompasses all changes in the visual characteristics of scene elements. Vegetation, vehicles, and buildings are the most sensitive to these shifts, as their appearance can vary due to seasonal changes, geographical differences, and even the time of day.

\begin{figure}[h!]
    \centering
    \includegraphics[width=\linewidth,trim={1cm 1cm 4cm 1.6cm},clip]{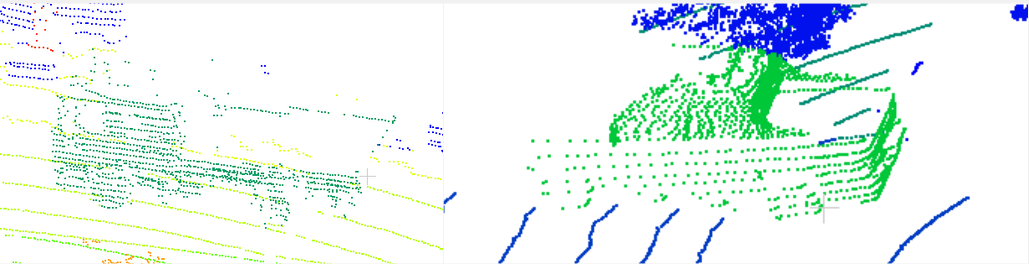}
    \caption{Illustration of appearance shift: On the left, an object labeled as a truck from the SemanticKITTI dataset~\cite{semantickitti} in Germany, and on the right, an object labeled as a truck from the nuScenes dataset~\cite{nuscenes} in the US.}
    \label{fig:app_change}
\end{figure}

In~\autoref{fig:app_change}, the appearance shift is illustrated using the example of trucks from Germany and the US. Although both serve the same purpose—carrying large objects in their trunks—they have significantly different visual appearances.

\paragraph{Scene shift} it encompasses two types of changes related to variations in scene composition. First, it involves changes in the types of objects expected in different environments. For example, traffic lights are common in urban areas but rare on freeways. Second, it reflects changes in the behavior of road users, which can affect the location and quantity of various elements within the scene.

\begin{figure}[h!]
    \centering
    \includegraphics[width=\linewidth,trim={3cm 4cm 3cm 3cm},clip]{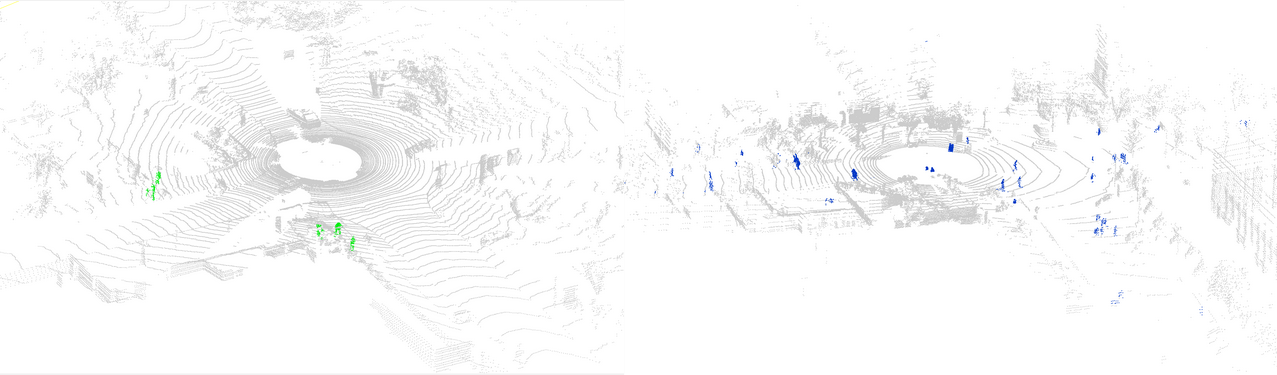}
    \caption{Illustration of scene shift: On the left, a scan from the SemanticKITTI dataset~\cite{semantickitti} in a German suburban area, and on the right, a scan from the SemanticPOSS dataset~\cite{semanticposs} on a university campus. Pedestrians are highlighted in green on the left, where they are located on sidewalks, and in blue on the right, where they are dispersed throughout the scene.}
    \label{fig:sc_change}
\end{figure}

This second point is illustrated in~\autoref{fig:sc_change}, where pedestrians are highlighted. In a campus scene, pedestrians are scattered almost randomly throughout the area, whereas in a suburban scene, they are mostly confined to sidewalks.

\paragraph{Sensor shift} Although sensor shift occurs for cameras, due to factors like focal length and exposure time, it is far more pronounced for LiDAR sensors.

Sensor shift encompasses all sensor-related sources of domain variation, including intrinsic properties such as sensor technology (rotating vs. solid-state), vertical and angular resolution, and field of view, as well as extrinsic factors like the sensor's placement on the acquisition vehicle.

Sensor shift is further amplified by the lack of consensus among dataset providers on the optimal sensor setup and model. Consequently, sensor shifts can be observed between each pair of datasets.

\begin{figure}[h!]
    \centering
    \includegraphics[width=\linewidth,trim={1cm 1cm 1cm 1cm},clip]{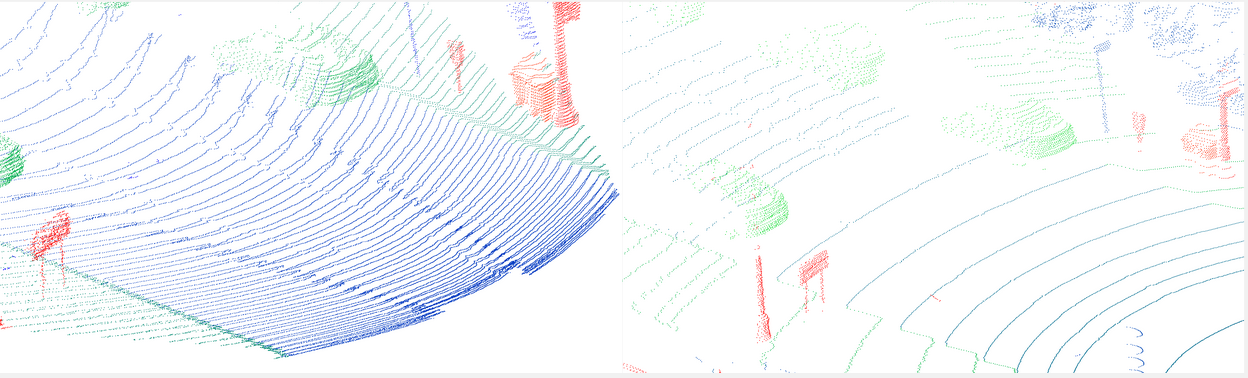}
    \caption{Illustration of sensor shift: Both scans were acquired simultaneously from the PandaSet dataset~\cite{pandaset}. On the left is a scan from a solid-state LiDAR, and on the right is a scan from a 64-beam rotating LiDAR.}
    \label{fig:se_change}
\end{figure}

We illustrate sensor shift in~\autoref{fig:se_change}. This figure shows two synchronized acquisitions of the same scene from the PandaSet dataset~\cite{pandaset} using different sensors. Despite the absence of scene and appearance shifts, the resulting scans appear noticeably different.

\subsection{Datasets}

\begin{table*}[!ht]
    \centering
    \begin{tabular}{l|cccccc}
        Dataset & Year &  Abbreviation & Country & Scene type & LiDAR sensor & \# labels  \\ 
        \hline
        SemanticKITTI~\cite{semantickitti} & 2019 & SK & Germany & suburban & HDL-64E & 19\\
        nuScenes~\cite{nuscenes} & 2020 & NS & US+Singapore & urban & HDL-32E & 16\\
        Waymo~\cite{waymo} & 2020 & W & US & urban & 64-beam rotating & 22\\
        SemanticPOSS~\cite{semanticposs} & 2020 & SP & China & campus & Pandora & 13\\
        PandaSet~\cite{pandaset} & 2021 & PS (P64+PFF) & US & suburban & Pandar64+PandarGT & 37\\
        ParisLuco3D~\cite{parisluco3d} & 2024 & PL3D & France & urban & HDL-32E & 45\\
    \end{tabular}
    \vspace{2mm}
    \caption{Information on the datasets used in this study of domain generalization in LiDAR Semantic Segmentation. P64 refers to the scans captured by the Pandar64 LiDAR sensor (64-beam rotating) in PandaSet, while PFF refers to the scans from the PandarGT LiDAR sensor (front-facing solid-state).}
    \label{tab:summary_dataset}
\end{table*}

Over the past few years, numerous 3D semantic segmentation datasets in the field of autonomous driving have been released, leading to a wide range of setups and, consequently, a diverse array of domains to explore.

While synthetic datasets from simulators like GTA V~\cite{squeezeseg} or CARLA~\cite{CARLA, kitticarla} offer several advantages, particularly in terms of labeling as they are not susceptible to label shift~\cite{transfer}, the synthetic-to-real domain gap remains unsolved~\cite{cosmix} and is beyond the scope of this research.  Here, we focus on real-world datasets, specifically: SemanticKITTI~\cite{semantickitti}, nuScenes~\cite{nuscenes}, Waymo~\cite{waymo}, SemanticPOSS~\cite{semanticposs}, PandaSet~\cite{pandaset}, and ParisLuco3D~\cite{parisluco3d}. 

We further divide these datasets into two groups: training datasets (SemanticKITTI, nuScenes) and evaluation datasets (SemanticPOSS, PandaSet, Waymo, ParisLuco3D). The training datasets were selected due to their size and prominence in the literature, as both SemanticKITTI and nuScenes are commonly used for benchmarking segmentation methods. In addition, we use SemanticKITTI-32, a subsampled version of SemanticKITTI. This dataset is valuable for assessing sensitivity to sensor shift, specifically changes in acquisition resolution, without introducing appearance or scene shifts. Previous domain generalization studies focused on a smaller set of datasets, leading to a more limited exploration of domain shifts.

PandaSet~\cite{pandaset} is a unique dataset as it was collected using two synchronized LiDAR sensors with different technologies: a rotating LiDAR and a solid-state LiDAR. Throughout the rest of this work, these acquisitions will be treated separately and referred to as Panda64 (P64) and PandaFF (PFF), respectively.

%As we interest ourselves in domain generalization, it is interesting to highlight the high variability of information represented in each datasets. In the \autoref{tab:comp_sensor}, we compare the sequence properties of each dataset. We observe a large variety of acquisition sensor, 5 different for 6 datasets. 

In~\autoref{tab:summary_dataset}, we provide a summary of the metadata for the various datasets. The table highlights the diversity of acquisition locations and scene types, which will be essential for studying domain generalization.

%\begin{table}[h!]
%\scalebox{0.9}{
%    \centering
%    \begin{tabular}{l|ccc}
%        \multicolumn{1}{c}{Dataset} & \multicolumn{1}{c}{Length (m)} & \multicolumn{1}{c}{Duration (s)}  &  \multicolumn{1}{c}{\tabcell{Angular resolution (°) \\ Horizontal/vertical}}\\ \midrule \hline
%        SemanticKITTI \cite{semantickitti}& 2015 &209 & 0,08/0,4\\ \hline
%        nuScenes \cite{nuscenes}& 100&20 & 0,33/1,33\\ \hline
%        Waymo \cite{waymo}& 103&20 & 0,16/0,31\\ \hline
%        SemanticPOSS \cite{semanticposs}&230 &50 & 0,2/de 0,33 à 1\\ \hline
%        Panda64 \cite{pandaset}& 82 & 8& 0,2/from 0,17 to 5 \\ \hline
%        ParisLuco3D~\cite{parisluco3d} & 2135 & 750 & 0,16/1,33\\ \hline \midrule
%    \end{tabular}
%}
%    \caption{Comparison of the acquisition average properties of the various datasets' sequences.}
%    \label{tab:comp_sensor}
%\end{table}

\subsection{Label sets}

Cross-dataset evaluation is challenging because each dataset has a distinct label set. Additionally, as noted in~\cite{parisluco3d}, label shift must be considered to ensure a fair evaluation. With this in mind, we have created nine separate label sets tailored for evaluation. These label sets correspond to the different combinations of training and evaluation datasets and are as follows:

\begin{itemize}
    \item $\mathcal{L}_{SK\cap SP}$: \textit{person, rider, bike, car, ground, trunk, vegetation, traffic sign, pole, building, fence}
    \item $\mathcal{L}_{SK\cap PS}$: \textit{2-wheeled, pedestrian, driveable ground, sidewalk, other ground, manmade, vegetation, 4-wheeled}
    \item $\mathcal{L}_{SK\cap NS}$: \textit{motorcycle, bicycle, person, driveable ground, sidewalk, other ground, manmade, vegetation, vehicle, terrain}
    \item $\mathcal{L}_{SK\cap W}$: \textit{car, bicycle, motorcycle, truck, vegetation, sidewalk, road, person, bicyclist, motorcyclist, trunk, other-vehicle, sign, pole, building, other-ground}
    \item $\mathcal{L}_{SK\cap PL3D} = \mathcal{L}_{SK}$
\end{itemize} 
\vspace{0.2cm}
\begin{itemize}
    \item $\mathcal{L}_{NS\cap SP}$: \textit{person, bike, car, ground, vegetation, manmade}
    \item $\mathcal{L}_{NS\cap PS}$: \textit{2-wheeled, pedestrian, driveable ground, sidewalk, other ground, manmade, vegetation, 4-wheeled}
    \item $\mathcal{L}_{NS\cap W}$: \textit{car, truck, bus, other vehicle, motorcycle, bicycle, pedestrian, traffic cone, manmade, vegetation, driveable road, sidewalk, other ground}
    \item $\mathcal{L}_{NS\cap PL3D} = \mathcal{L}_{NS}$
\end{itemize}

Although the use of multiple label sets complicates direct comparison of results compared to using a single common label set, as done in~\cite{domgen}, we believe this approach allows for more detailed and accurate insights. Relying on just one global label set can obscure important distinctions present in certain datasets, such as the differentiation between bicyclists and motorcyclists. We will also use the terms $\mathcal{L}_{SK}$ and $\mathcal{L}_{NS}$ to refer to the standard label sets for SemanticKITTI and nuScenes, respectively. 

The ParisLuco3D~\cite{parisluco3d} dataset was annotated to include all labels from both the SemanticKITTI and nuScenes datasets, specifically designed for cross-dataset evaluation. This is why all SemanticKITTI and nuScenes labels are present in ParisLuco3D, allowing for direct evaluation without the need to merge classes.

To evaluate domain generalization performance, we will use the per-class intersection-over-union (IoU) and the mean intersection-over-union (mIoU). Since the mIoU depends on the label set used, we will clarify the label set and the evaluation set by using the following notation: mIoU$^{\mathrm{evaluation-set}}_{\mathrm{label-set}}$. For example, when evaluating domain generalization from a model trained on SemanticKITTI and tested on PandaSet with the Pandar64 scans, we will calculate the mIoU using the label set $\mathcal{L}_{SK\cap PS}$ and denote the result as mIoU$_{\mathcal{L}_{SK\cap PS}}^{P64}$, simplified as P64 in some tables.

\section{3DLabelProp}

\subsection{Motivation}

As noted in the related work section, recent LiDAR domain generalization methods have concentrated on addressing the sensitivity to shifts in 3D sensor data. Following the same approach, we aim to identify a canonical domain where sensor shift is minimal or nearly absent. Complete \& Labels~\cite{completelabel} (C\&L) employed a learned scene completion model to define this canonical domain. While this concept is interesting, in the context of domain generalization, it simply shifts the requirement for robustness from semantic segmentation to scene completion. We consider this approach inadequate for consistently identifying a canonical domain.

In this work, we propose using pseudo-dense point clouds by leveraging the sequential nature of autonomous driving datasets. These pseudo-dense point clouds are created by performing LiDAR odometry and combining multiple consecutive LiDAR scans, resulting in locally dense point clouds that are expected to be less affected by sensor shifts. This approach assumes that the 3D registration process remains robust to sensor shifts; otherwise, we would encounter similar limitations to those in C\&L. LiDAR SLAM has demonstrated robustness to sensor variations in autonomous driving applications~\cite{ct-icp}, helping to mitigate this issue.

In~\autoref{fig:teaser}, we illustrate pseudo-dense point clouds. In this pseudo-dense domain, the road and buildings appear almost identical, whereas they showed notable differences in the single-scan domain due to the significant variation in sensor topology.

%\begin{figure*}[h!]
%    \centering
%    \subfloat{
%        \includegraphics[width=.5\linewidth]{images/sk_scak.png}
%    }
%    \subfloat{
%        \includegraphics[width=.5\linewidth]{images/sk32_scak.png}
%    } 
%    \\
%    \centering
%    \subfloat{
%        \includegraphics[width=.5\linewidth]{images/sk_acc.png}
%    }
%    \subfloat{
%        \includegraphics[width=.5\linewidth]{images/sk32_acc.png}
%    } 
%    \caption{Comparisons of scene in the single-scan and the pseudo-dense domains. The original scene are from SemanticKITTI (left) and SemanticKITTI-32 (right).}
%    \label{fig:pseudo-dense}
%\end{figure*}

Although we anticipate that operating in the pseudo-dense domain will help mitigate sensor shift and thereby enhance domain generalization performance, this approach has its drawbacks. Pseudo-dense methods produce significantly larger data than single LiDAR scans, leading to longer processing times. To address this, acceleration strategies are essential to improve processing efficiency.

Our domain generalization method, 3DLabelProp, utilizes pseudo-dense point clouds while integrating several geometric modules, specifically label propagation and clustering, to accelerate processing while preserving all dense information.

\begin{table*}[ht]
\footnotesize
    \centering
    \begin{tabular}{l|c|ccccccc}
       \textbf{Method} & \tiny mIoU$_{\mathcal{L}_{SK}}^{SK}$ & \tiny mIoU$_{\mathcal{L}_{SK}}^{SK32}$ & \tiny mIoU$_{\mathcal{L}_{SK\cap PS}}^{P64}$ & \tiny mIoU$_{\mathcal{L}_{SK\cap PS}}^{PFF}$  & \tiny mIoU$_{\mathcal{L}_{SK\cap SP}}^{SP}$ & \tiny mIoU$_{\mathcal{L}_{SK\cap W}}^{W}$ & \tiny mIoU$_{\mathcal{L}_{SK\cap NS}}^{NS}$  & \tiny mIoU$_{\mathcal{L}_{SK\cap PL3D}}^{PL3D}$  \\[1mm]
       \hline   
        KPConv~\cite{kpconv}  &58,3& 52,7 & 32,7 & 21,1& 39,1& 17,5 & 46,7 & 22,1 \\
        KPConv pseudo-dense~\cite{kpconv}  & 53,6 & 53,2 & 47,8  & \textbf{45,0} & 47,5& \textbf{N/A} &46,1  & \textbf{30,5} \\
        SRU-Net~\cite{mink} & 58,6 & 54,0 & 44,2 & 22,2& 45,3& 33,1 & 42,7 & 33,3 \\
        SRU-Net pseudo-dense~\cite{mink} & 56,7& 57,3 & \textbf{N/A} & \textbf{61,7} & 46,9 & 25,6 & 49,8 & \textbf{38,0} \\
        \end{tabular}
        \vspace{2mm}
        \caption{Comparison of domain generalization performance between models trained on single LiDAR scans and those trained on pseudo-dense point clouds. All models were trained using only the SemanticKITTI dataset. \textbf{N/A} indicates results unavailable due to memory limitations in handling pseudo-dense point clouds. In bold, the results on the most challenging datasets for domain generalization.}
    \label{tab:accumu}
\end{table*}

\subsection{Pseudo-dense point clouds}

To investigate pseudo-dense point clouds, we center our analysis on two semantic segmentation methods: SRU-Net~\cite{mink} and KPConv~\cite{kpconv}. These methods are standard for semantic segmentation of LiDAR scans and dense point clouds, respectively. SRU-Net has also been used as the backbone for other domain generalization approaches, including DGLSS~\cite{domgen} and LiDOG~\cite{lidog}.

To test our initial hypothesis, that pseudo-dense point clouds would enhance domain generalization performance, we train both models on either single LiDAR scans or pseudo-dense point clouds. The pseudo-dense point clouds are generated by combining the previous 20 scans with the current LiDAR scan using CT-ICP~\cite{ct-icp}, a robust LiDAR SLAM technique. 

The quantitative comparison of domain generalization performance for the trained models is presented in~\autoref{tab:accumu}. Due to data format limitations, KPConv with pseudo-dense data could not be tested on Waymo. Additionally, SRU-Net with pseudo-dense data was unable to perform inference on Panda64, as it consistently exceeded available computing resources (Nvidia RTX 3090), highlighting another challenge of pseudo-dense point clouds: high memory consumption.

Firstly, we observe a significant improvement in domain generalization performance in most cases. Notably, there is no performance drop for models trained on pseudo-dense point clouds when evaluated on SemanticKITTI-32, compared to their performance on SemanticKITTI, supporting our hypothesis that the pseudo-dense domain is almost free from sensor shift. Additionally, pseudo-dense methods are capable of extracting meaningful information from PandaFF (PFF), while single-scan methods fail to do so. However, using pseudo-dense point clouds negatively impacts source-to-source performance, as pseudo-dense methods consistently perform worse than single-scan methods.

\begin{table}[h!]
    \centering
    \begin{tabular}{l|cc}
        \textbf{Method} & SemanticKITTI (Hz)  & nuScenes (Hz) \\
        \hline   
        KPConv~\cite{kpconv} & 0,6 & 1,3 \\
        KPConv pseudo-dense~\cite{kpconv}& 0,1& 0,3\\
        SRU-Net~\cite{mink} & 6,7 & 7,7 \\
        SRU-Net pseudo-dense~\cite{mink} & 1,1 & 4,0 \\
    \end{tabular}
    \vspace{2mm}
    \caption{Processing speed comparison of KPConv~\cite{kpconv} and SRU-Net~\cite{mink} on single LiDAR scans versus pseudo-dense point clouds across the nuScenes and SemanticKITTI datasets. 10Hz and 20Hz are the targets to achieve real-time methods on the SemanticKITTI and nuScenes datasets, respectively.}
    \label{tab:vitesse_exec}
\end{table}

In the previous subsection, we noted that pseudo-dense methods are expected to be slower due to the increased point count. In \autoref{tab:vitesse_exec}, we compare processing speeds on nuScenes and SemanticKITTI based on input type. The results show that processing speed ranges from good or decent with single scans to insufficient or poor with pseudo-dense inputs.

\begin{table}[h!]
    \centering
    \resizebox{1.0\linewidth}{!}{
    \begin{tabular}{l|c|ccccccc}
        \textbf{Method} & SK & SK32 & PFF  & SP & NS  & PL3D \\
        \hline   
        SRU-Net pseudo-dense~\cite{mink} & 56,7 & 57,3 & 61,7 & 46,9 & 49,8 & 38,0 \\
        Same with SLAM pertubations & 45,1 & 41,5 & 58,5 & 39,9 & 36,3 & 30,1 \\
        \end{tabular}
        }
        \vspace{2mm}
        \caption{Domain generalization results for SRU-Net with pseudo-dense input without and with SLAM perturbations. All models are trained on SemanticKITTI.}
    \label{tab:slam_noisy}
\end{table}

Although LiDAR methods are generally robust to sensor shifts, it is worthwhile to examine how pseudo-dense methods respond to SLAM failures. In \autoref{tab:slam_noisy}, we assess SRU-Net with pseudo-dense input using artificially noisy SLAM positions to simulate poor trajectory estimation. The added noise is significantly higher than typical levels to mimic a failing SLAM. We observe a consistent performance decline, even compared to single-scan results, with the effect being more pronounced for lower-resolution sensors (nuScenes and ParisLuco3D). Throughout the remainder of this study, no failing SLAM trajectories were observed in any tested dataset.

In conclusion, pseudo-dense methods achieve satisfactory domain generalization performance but exhibit subpar source-to-source performance and slower processing speeds. 3DLabelProp is designed to address these limitations.

\subsection{3DLabelProp algorithm}

3DLabelProp is inspired by 2D video semantic segmentation techniques, which differentiate between two types of frames: simple frames, easily segmented using optical flow, and complex frames that require processing by the semantic segmentation model.

Similarly, for pseudo-dense point clouds, we differentiate between two types of points: simple points, which can be segmented geometrically, and complex points, which require processing by a learning model. Geometric methods are much faster than learning models, so we aim to minimize the region that the learning model needs to process.

Intuitively, simple points correspond to static objects. Since these objects remain stationary across frames in the global reference frame, we can utilize past data to identify new samples of these objects. The assumption that static objects have been previously sampled is known as the 4D-neighbor hypothesis, which posits that newly sampled points have 4D neighbors from the same object. For these points, we can then apply a nearest-neighbor-based propagation from previous frames to the current frame.

\begin{figure}[h!]
\centering
    \includegraphics[width=0.49\linewidth,trim={1cm 1cm 1cm 1cm},clip]{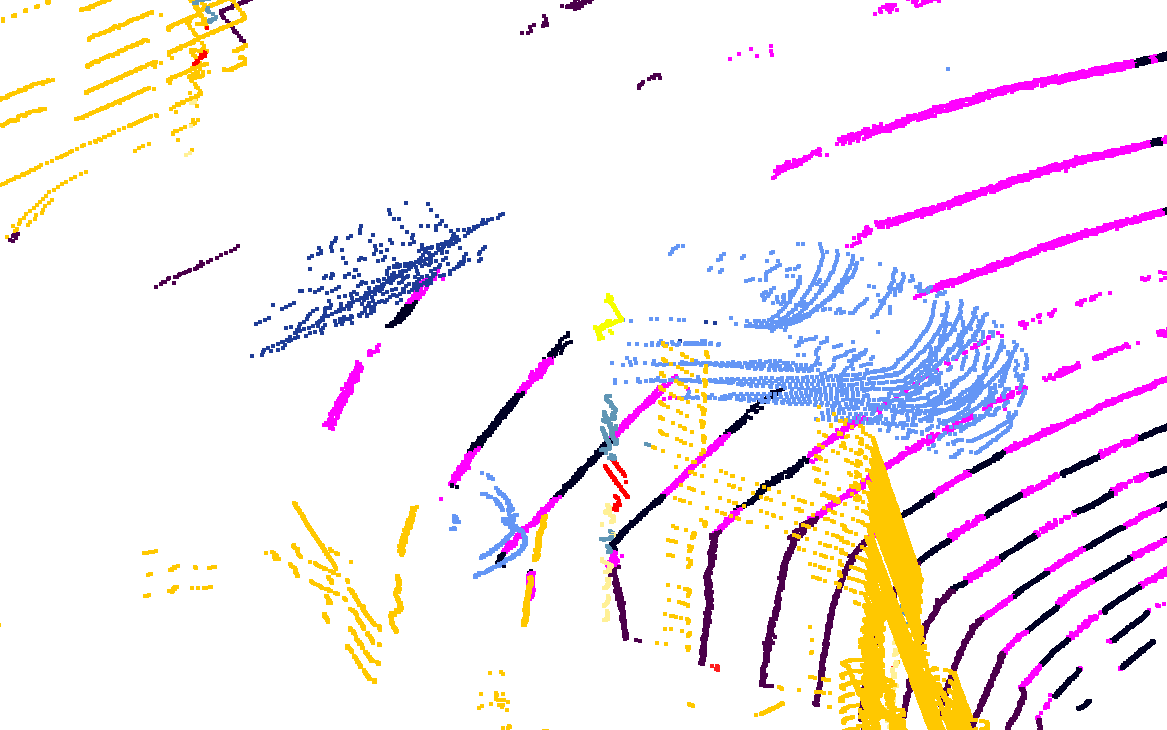}
    \includegraphics[width=0.49\linewidth,trim={1cm 1cm 1cm 1cm},clip]{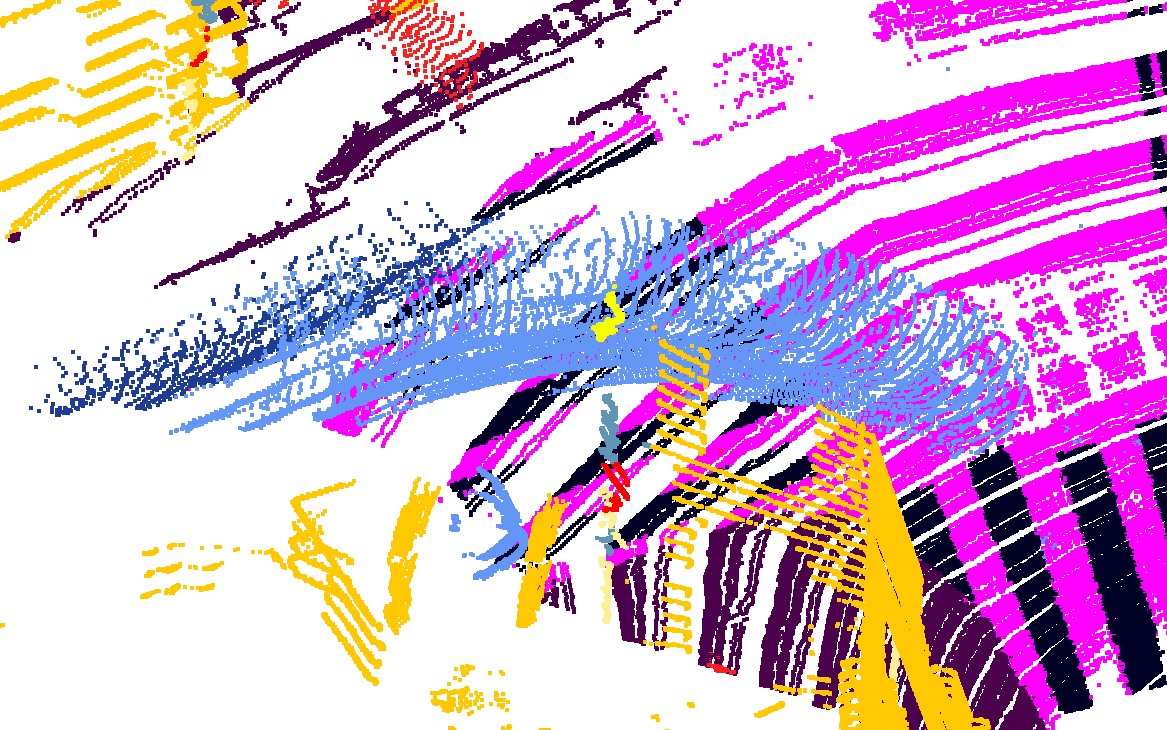}
    \caption{Illustration of the trail phenomenon. On the left, a section of a point cloud made up of 5 consecutive scans; on the right, a section composed of 20 consecutive scans.}
    \label{fig:trainees}
\end{figure}

Complex points correspond to dynamic objects, which move within the global reference frame, as well as newly sampled objects. These points lack meaningful neighbors in the pseudo-dense point clouds and therefore require processing by a learning model. It is crucial not to apply the 4D-neighbor hypothesis to moving objects, as their neighbors from previous frames in the global reference frame may belong to different objects due to the trail phenomenon, illustrated in~\autoref{fig:trainees}.

\begin{figure*}[h!]
    \centering
    \includegraphics[width=\linewidth]{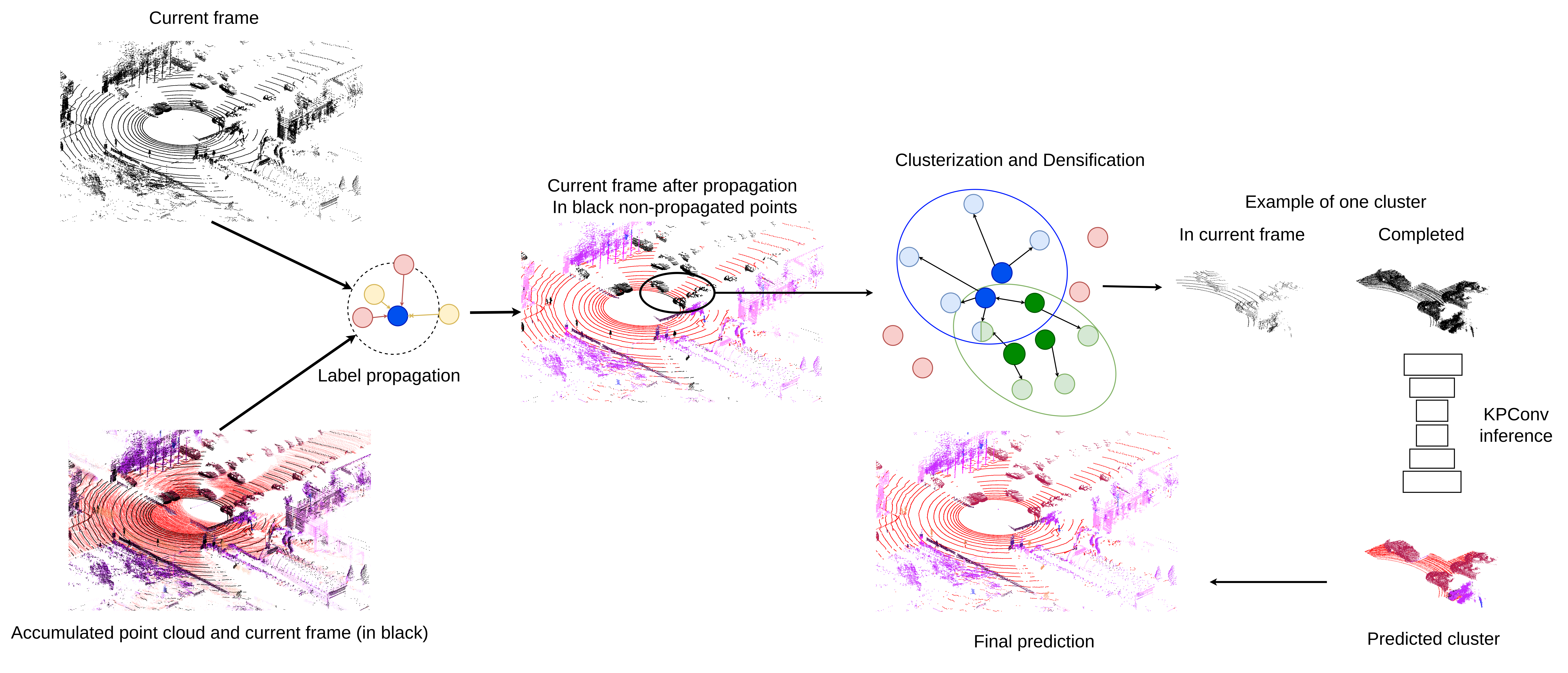}
    \caption{Overview of the 3DLabelProp method. Points from the current scan are accumulated with points from previous scans. A geometric propagation labels a large portion of the new points. The remaining points are grouped into clusters, densified with points from previous scans, and their semantics are inferred by a deep network for dense point clouds, KPConv. The predictions are then merged with the geometric labels to produce the final result.}
    \label{fig:3DLabelProp}
\end{figure*}

The 3DLabelProp method (illustrated in~\autoref{fig:3DLabelProp}) depends on registering the newly acquired LiDAR scan with the pseudo-dense point cloud from previous scans, known as the reference point cloud. This initial registration step is not covered in this work: we consistently use the LiDAR SLAM method CT-ICP~\cite{ct-icp}. The reference point cloud comprises the previous $N_s$ scans registered with CT-ICP. $N_s$ is a hyperparameter of the method, set by default to $N_s = 20$, and we demonstrate its impact on the method's accuracy and speed in the ablation study.

The 3DLabelProp method is then divided into five steps:
\begin{enumerate}
\item \textit{Propagation of Labels}: Propagate the labels of previously segmented static objects to new samples of these objects.
\item \textit{Clusterization}: Identify complex regions within the point cloud and divide them into \( K_c \) clusters.
\item \textit{Cluster densification}: Densify the clusters using 4D neighbors from the reference point cloud.
\item \textit{Cluster segmentation}: Segment the densified clusters using a learning model.
\item \textit{Prediction fusion}: Fuse predictions from both the geometric and learning models.
\end{enumerate}

An aspect not explicitly covered in the previous algorithm is memory footprint reduction. Although not algorithmically significant, this step is essential to address previously identified issues in processing pseudo-dense point clouds. The reference point cloud is sub-sampled using a $5~cm$ grid, keeping one point per voxel, and points beyond the acquisition range (set to $75~m$ here) are discarded. This is done concurrently with the LiDAR odometry.

We will explain the five steps in the following sections.

\subsubsection{Propagation of Labels}

First, it is essential to divide the label set, \( L = (l_i)_{i \in \{1, K\}} \), into two subsets: one for dynamic objects, \( D = (d_i)_{i \in \{1, K_d\}} \), and one for static objects, \( S = (s_i)_{i \in \{K_d+1, K\}} \). These subsets do not intersect, and \( L = D \cup S \).

Static objects are defined as immovable items, such as the ground and buildings. Dynamic objects include both moving and potentially movable items, such as pedestrians and vehicles. By definition, static objects remain stationary in the global reference frame from one scan to the next. Therefore, by defining a sufficiently small neighborhood, we can assume that two neighboring points are sampling the same object.

%In practice, we use a 10~cm radius and assign a newly sampled point the majority class within this neighborhood. Using ground truth label data, this step achieves 95\% accuracy for static objects on SemanticKITTI.

\begin{figure}[h!]
    \centering
    \includegraphics[width=0.49\linewidth,trim={2cm 14cm 1cm 0cm},clip]{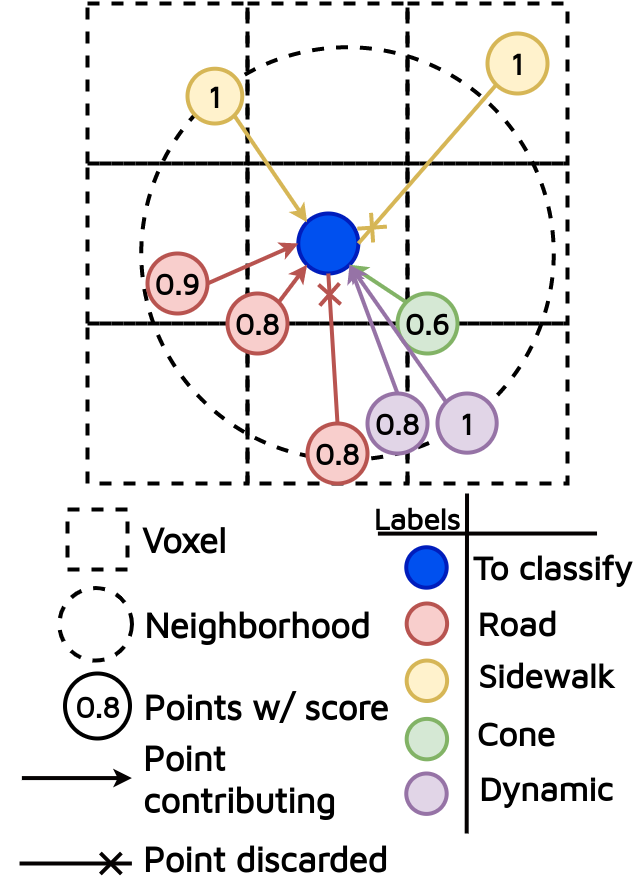}
    \includegraphics[width=0.49\linewidth,trim={2cm 0cm 0cm 17.5cm},clip]{images/lab_prop.png}
    \caption{Propagation module of our 3DLabelProp method. We use the segmentation scores of points from previous scans to label a new point in the current LiDAR scan.}
    \label{fig:prop}
\end{figure}

Two steps are required to complete the propagation: extracting the neighborhood for each point, and assigning labels and scores based on that neighborhood. The neighborhood is extracted through a radius search, accelerated by voxelizing the reference cloud (in all experiments, we used a voxel of $0.80~m$). Voxelization allows for pre-neighborhood extraction in \( O(1) \) time. The label is then determined by a voting process, with each vote weighted by the segmentation score and distance to the target point. If a vote's weight is too low, it is discarded, as the neighbor is deemed unreliable. If the voted label corresponds to a dynamic label, no label is assigned, as propagating dynamic objects is assumed to be unfeasible. \autoref{fig:prop} illustrates the propagation module.

Formally, let us introduce \( \mathcal{P}_{r} \in \mathbb{R}^{N \times 3} \), the reference point cloud composed of previous LiDAR scans, \( Y_{r} \in \{1, \dots, K\}^N \), the label for each point, and \( \mathcal{C}_{r} = (c_{i} \times e_{i})_{i \in \{1, \dots, N\}} \), the segmentation score for each point, represented as a one-hot encoded vector (\( e_{i} \)) multiplied by the confidence score for the associated label (\( c_{i} \)). Similarly, we define \( \mathcal{P}_{c} \in \mathbb{R}^{M \times 3} \), \( Y_{c} \), and \( \mathcal{C}_{c} \) as the corresponding values for the newly acquired LiDAR scan. Initially, the labels are set to -1 and the confidence scores to 0. We can then express the propagation as follows:

\begin{equation}
\forall p_{j} \in \mathcal{P}_{c}, \quad c_{j} = \sum_{p_{i} \in \mathcal{N}_{\mathcal{P}_{r}}(p_{j})}  w(i,j)    \mathds{1}_{w(i,j)>0,5};
\end{equation}

\noindent where \( w(i,j) = d(p_{i}, p_{j}) \times c_{i} \) and \( d(p, q) = e^{-\frac{||p - q||^2}{d_p^2}} \), with \( d_p \) as a hyperparameter of the method that determines the significance of the neighborhood in label propagation. By default, \( d_p \) is set as $0.30~m$. \( \mathcal{N}_{\mathcal{P}_{r}}(p_{j}) \) represents the neighborhood of point \( p_{j} \) from the current scan’s point cloud \( \mathcal{P}_{c} \) within the reference point cloud \( \mathcal{P}_{r} \) from previous scans.

We then have:
\[
        y_{j} = 
\begin{cases}
    \arg\max(c_{j}), & \text{if } \arg\max(c_{j}) > K_d \\
    -1, & \text{otherwise}
\end{cases}
\]
\noindent
where \(\arg\max(c_{j}) > K_d\) indicates that the label is static (with \(K_d\) representing the number of dynamic labels).

\begin{figure}[h!]
\centering
    %\includegraphics[width=.45\linewidth,trim={8cm 1cm 14cm 1cm},clip]{images/input_prop.png}
    %\hspace{1mm}
    %\includegraphics[width=.45\linewidth,trim={6cm 1cm 14cm 1cm},clip]{images/output_prop.png}
    \includegraphics[width=.9\linewidth,trim={6cm 4cm 10cm 2cm},clip]{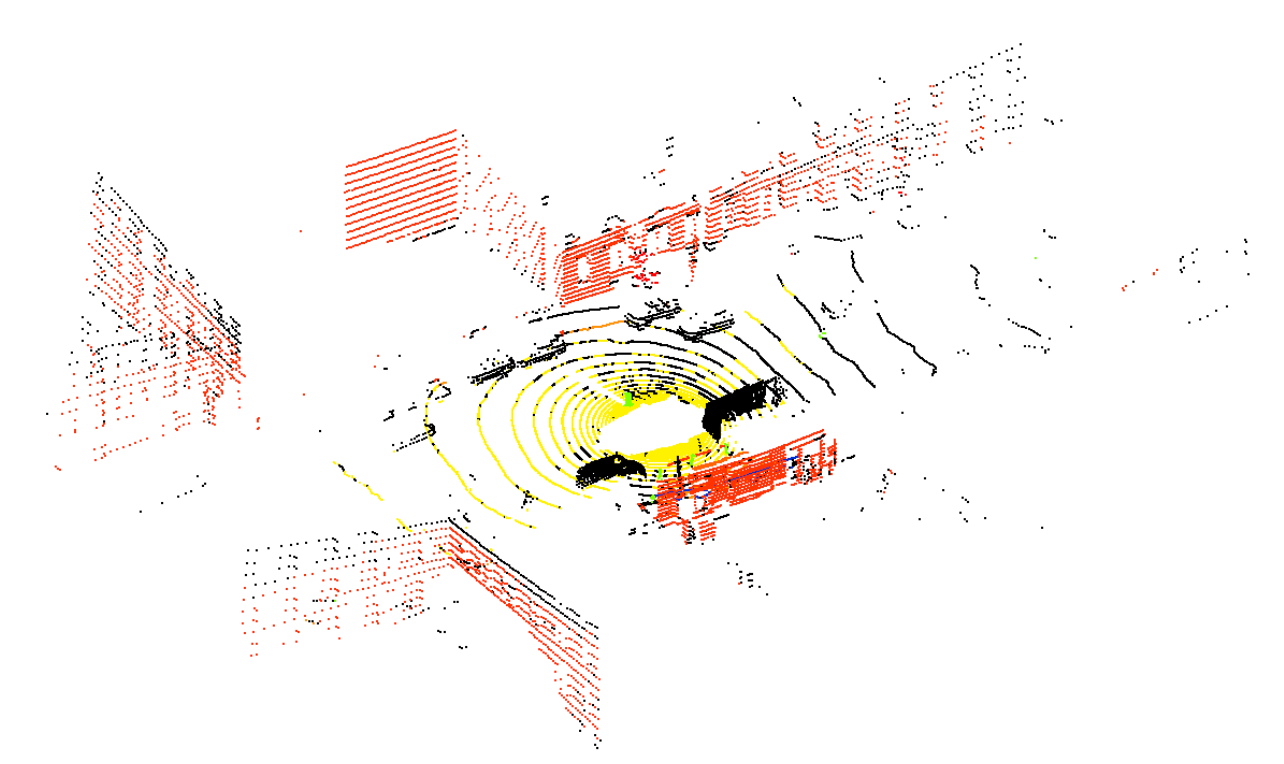}
    \caption{Results of the propagation module. In black, the points considered dynamic; in other colors, all points with propagated labels from previous scans.}
    \label{fig:prop-gt}
\end{figure}

The results of the propagation step are illustrated in~\autoref{fig:prop-gt}. 

%For visualization purposes, ground truth labels from previous scans were used; however, during inference, the labels from previous segmentations are applied.

\subsubsection{Clusterization}

In 3DLabelProp, we propose using a clustering algorithm to partition the point cloud more efficiently. This is applied immediately after the propagation module on the residual points, i.e., points \( p_{j} \) where \( y_{j} = -1 \). This subset is considerably smaller than the original point cloud (see~\autoref{fig:prop-gt}), allowing the use of more advanced clustering algorithms without significant computational overhead. We use K-means to maintain a consistent number of \( K_c \) clusters (\( K_c \) being a hyperparameter in our method, set by default to 20 clusters).

\subsubsection{Cluster densification}

\begin{figure}
    \centering
    \includegraphics[width=0.45\linewidth,trim={0cm 12.5cm 1cm 0cm},clip]{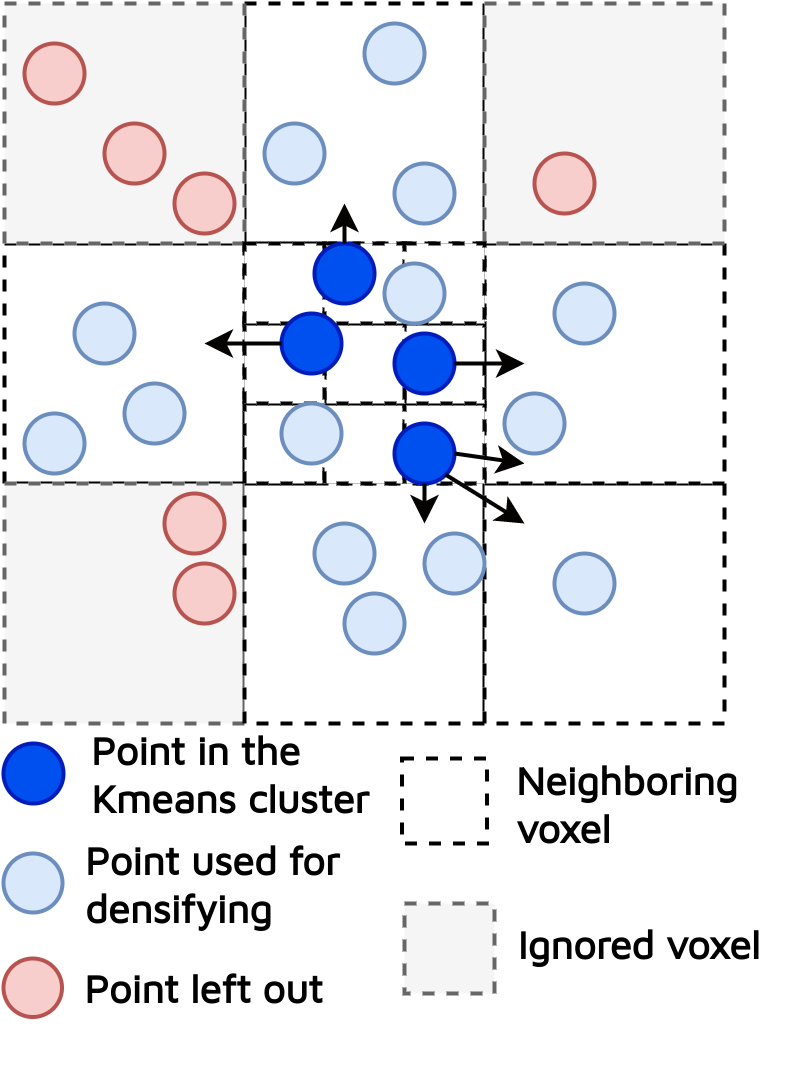}
    \includegraphics[width=0.45\linewidth,trim={0cm 2cm 1cm 26cm},clip]{images/cluster.png}
    \caption{Cluster densification module. We subdivide the voxels occupied by the original cluster into 27 (3x3x3) sub-voxels, each linked to a specific neighborhood. Only the neighbors of these occupied sub-voxels are included in the densification process.}
    \label{fig:densification}
\end{figure}

The extracted clusters are computed from the residual points in the current scans and therefore lack substantial contextual information. To address this, they are densified with points from the reference point cloud. This is done by voxelizing the point cloud and adding points from the reference point cloud within the same voxels, as well as their neighbors, to the cluster's contextual data. However, because the voxel size is relatively large to save processing time (we used $2~m$-sized voxels in all our experiments), some neighboring voxels could add many points with minimal contextual benefit. To optimize this, we divide the voxels occupied by the original cluster into 27 (3x3x3) sub-voxels, associating each with a specific neighborhood. Only the neighbors of the occupied sub-voxels are included in the densification process. The process is illustrated in 2D in \autoref{fig:densification}.

\begin{figure}
    \centering
    \includegraphics[width=.48\linewidth,trim={3cm 5cm 0cm 1.5cm},clip]{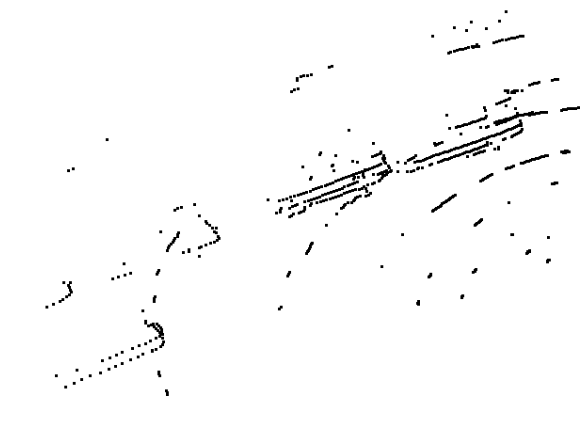}
    \hspace{0.5mm}
    \includegraphics[width=.48\linewidth,trim={8cm 8cm 0cm 1.5cm},clip]{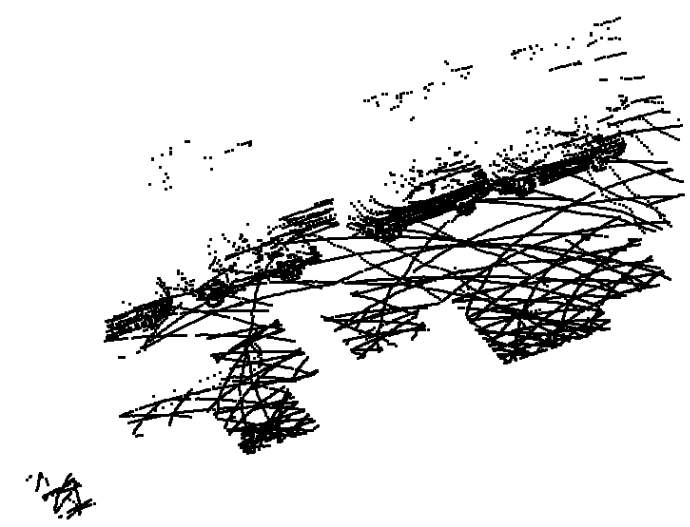}
    \caption{On the left, an extracted cluster from the residual points in the current scan; on the right, the same cluster after densification with points from previous scans.}
    \label{fig:cluster}
\end{figure}

An illustration of the densification process on a cluster in 3D taken from a scan is shown in~\autoref{fig:cluster}.

\subsubsection{Cluster segmentation}

Once the clusters are densified, they are processed with a deep learning algorithm that ideally leverages dense neighborhoods. Since these clusters do not follow the typical acquisition pattern of rotating LiDAR, range-based methods are not applicable. In this work, we use KPConv~\cite{kpconv}, which achieves state-of-the-art performance on dense point clouds and has an implementation that enables efficient parallel processing of clusters. KPConv is specifically re-trained on densified clusters, with labels propagated using ground truth during training. At inference time, no adaptation is applied; label propagation relies on past inferences.

\subsubsection{Prediction fusion}

To generate the final segmented point cloud, we must merge predictions from both KPConv and the label propagation module. During cluster densification, contextual points inferred via the label propagation module may be included, resulting in some points having two potential predictions, as KPConv infers a semantic to all points within a cluster. We consistently retain the label predictions from the KPConv deep network, as it is more reliable than the geometric propagation prediction.

%To address this, we introduce a hyperparameter \( \alpha \) to balance the influence of KPConv and the label propagation module. This produces a weighted score per class, and we use the argmax to select the final predicted class.

\subsection{Training protocol of KPConv}

KPConv model used inside 3DLabelProp is the standard layout, namely the KPFCNN. It consists of 4 downsampling blocks and 4 upsampling blocks. In every case, the model architecture does not change. For the training, we use a Lovasz loss and weighted cross-entropy loss. We use a SGD optimizer with a weight decay of 1e-4 and a momentum of 0.98. The learning rate scheduler is a cosine annealing scheduler. It is trained with cluster extracted by the 3DLabelProp pipeline when assuming that past inferences are the ground truth. Training hyperparameters are: learning rate: 0.005, batch size: 12, number of iterations: 400 000 (for SemanticKITTI); learning rate: 0.001, batch size: 16, number of iterations: 350 000 (for nuScenes).

\section{Domain generalization benchmark of LiDAR Semantic Segmentation (LSS) methods and results of our 3DLabelProp approach}

\subsection{Description of experiments}

Most current domain generalization methods concentrate on a single semantic segmentation model and demonstrate their method’s effectiveness by comparing it to baselines computed using that model. To our knowledge, no research has specifically examined the generalization performance of different 3D neural network models in real-to-real scenarios. In~\cite{robustness}, a robustness benchmark is conducted on various semantic segmentation models; however, this is limited exclusively to synthetic evaluation datasets built from SemanticKITTI.

%A tedious work of grouping together baselines from domain generalization and domain adaptation works could be done to extract a naive domain generalization benchmark. This work could be done under the assumption that every results extracted would be computed on the same label set, which is not the case.

In this section, we provide an overview of the domain generalization performance of current state-of-the-art LiDAR semantic segmentation models and their sensitivity to various domain shifts. Once this benchmark is established, we can compare 3DLabelProp against it.

\subsubsection{Selection of models to test}

Several criteria were considered in selecting models for inclusion in the benchmark: source code availability, availability of pre-trained weights, the paradigms they follow, and their performance on single-source benchmarks (nuScenes and SemanticKITTI).

The selected models are: Cylinder3D~\cite{cylinder3d}, KPConv~\cite{kpconv}, SRU-Net~\cite{mink}, SPVCNN~\cite{spvnas}, Helix4D~\cite{helix4D}, and CENet~\cite{cenet}. 

These models encompass a variety of input representations (point-based, voxel-based, range-based, etc.) and all achieve satisfactory source-to-source performance.

\subsubsection{Selection of experiments}

To assess a model's suitability for domain generalization, it is crucial to test it under various types of domain shifts. For each experiment the model undergoes, understanding the specific domain shifts involved is essential for drawing accurate conclusions. As previously mentioned, SemanticKITTI and nuScenes are used as training sets due to their size. Using two distinct training sets helps ensure that observed trends are not specific to a particular sensor. Additionally, the differences between these datasets increase the variety of domain shifts encountered.

A summary of the experiments we will conduct and the conclusions we aim to draw from them is presented in~\autoref{tab:experiments}. In total, we propose 14 evaluation dataset experiments. It is important to clarify that each method was trained on the standard label set of the source data.

\begin{table}[h!]
    \centering
    \resizebox{1.0\linewidth}{!}{
    \begin{tabular}[h]{lcc}
       \textbf{Training set} & \textbf{Evaluation set} & \textbf{Primary shifts evaluated for DG} \\
       \hline
       SK & SK32 & Sensor shift\\
       SK & P64, and PFF & Sensor shift between P64 and PFF\\
       SK & SP, and W & Appearance+Scene shift\\
       SK & NS & All domain shifts\\
       SK & PL3D & All domain shifts without label shift \\
       \hline
       NS & SK, and SK32 & Sensor shift between SK and SK32 \\
       NS & P64, and PFF & Sensor shift between P64 and PFF\\
       NS & SP, and W & All domain shifts\\
       NS & PL3D & Appearance+Scene shifts without label shift \\
    \end{tabular}
    }
    \vspace{2mm}   
    \caption{Summary of our LiDAR domain generalization experiments (SK is SemanticKITTI and NS is nuScenes).}
    \label{tab:experiments}
\end{table}

In detail, we include two pairs of datasets (SemanticKITTI and SemanticKITTI-32, Panda64 and PandaFF) to analyze the effects of sensor shift. In both cases, the same scene is captured by two different sensors, ensuring no other domain shifts are present. SemanticKITTI and SemanticKITTI-32 use similar sensors, differing only in resolution, while Panda64 and PandaFF employ vastly different sensors, resulting in a much more pronounced sensor shift.

SemanticPOSS offers a unique type of scene, as it was captured on a university campus. Consequently, there is a significant scene shift compared to SemanticKITTI and nuScenes, with numerous pedestrians and bikes behaving quite differently from those in the training sets.

SemanticPOSS provides a particular type of scene as it is acquired on a university campus. As a result, there is a major scene shift relatively to SemanticKITTI and nuScenes. There are plenty of pedestrians and bikes, behaving in a much different manner than in the training sets.

Waymo introduces appearance and scene shifts with detailed annotations across a large number of sequences, resulting in a much more refined evaluation label set compared to SemanticPOSS.

Finally, ParisLuco3D offers an experimental setup without label shift, enabling analysis of per-class generalization results. Additionally, there is no sensor shift when training on nuScenes and testing on ParisLuco3D (they use both the same Velodyne HDL32 LiDAR sensor).

\subsection{Experimental protocol}

To build our benchmark, we re-trained each model on SemanticKITTI and nuScenes. To ensure performance comparable to that reported by the authors, we used their recommended hyperparameters. When no parameters were specified for nuScenes, we applied those used for SemanticKITTI. Re-training the models allows consistent use of the same data augmentations: rotation around the z-axis, scaling, and local Gaussian noise. Additionally, the models were trained without the reflectivity channel, which was replaced by occupancy. Reflectivity, a sensor-specific attribute for each LiDAR point, has been shown to hinder domain generalization, as noted in~\cite{vfieldlss}. To further support this claim, we examine the impact of reflectivity on domain generalization performance in the following section.

\subsection{Influence of reflectivity on the generalization of LSS models}

In~\autoref{tab:DG_R}, the selected models were trained on SemanticKITTI with and without LiDAR reflectivity and evaluated on SemanticKITTI and SemanticPOSS. SemanticPOSS was chosen for this analysis due to its angular resolution, which closely matches that of SemanticKITTI, resulting in minimal sensor shift. 

We can see that, consistently across all models, generalization improves when LiDAR reflectivity is not used. This is despite the fact that the reflectivity values in SemanticPOSS have a similar distribution to those in SemanticKITTI.

On the other hand, source-to-source performance consistently decreases when LiDAR reflectivity is removed.

\begin{table}[h!]
        \centering
        \resizebox{0.8\linewidth}{!}{
        \begin{tabular}{lccc}
           \textbf{Model} & Reflectivity & mIoU$_{\mathcal{L}_{SK}}^{SK}$ & mIoU$_{\mathcal{L}_{SK\cap SP}}^{SP}$ \\
           \noalign{\vskip 0.5mm}
           \hline
            \multirow{2}{*}{CENet \cite{cenet}} &\checkmark & \textbf{61,4} &27,5\\
            & $\times$&58,8&\textbf{27,9} \\ \cline{1-4}
            \multirow{2}{*}{Helix4D \cite{helix4D}} &\checkmark & \textbf{63,1} & 35,0 \\
            & $\times$& 60,0 & \textbf{36,0}\\ \cline{1-4}
            \multirow{2}{*}{KPConv \cite{kpconv}} & \checkmark & \textbf{59,9} &33,1 \\
            & $\times$ & 58,3 &\textbf{39,1}\\ \cline{1-4}
            \multirow{2}{*}{SRU-Net \cite{mink}} &\checkmark & \textbf{61,9} &45,2\\
            & $\times$&58,6 &\textbf{45,3}\\ \cline{1-4}
            \multirow{2}{*}{SPVCNN \cite{spvnas}} &\checkmark & \textbf{63,8} & 36,9\\
            & $\times$& 62,3 &\textbf{45,4} \\ \cline{1-4}
            \multirow{2}{*}{C3D \cite{cylinder3d}} & \checkmark& \textbf{66,9} & 33,8 \\ 
           & $\times$& 60,7&\textbf{41,9} \\ 
        \end{tabular}
        }
        \vspace{1mm}
        \caption{Source-to-source and domain generalization results with and without LiDAR reflectivity. All models were trained on SemanticKITTI (SK) and tested on SemanticPOSS (SP).}
        \label{tab:DG_R}
\end{table}

Since the decrease in source-to-source performance is undesirable, we explore an alternative strategy to handle reflectivity. We propose applying reflectivity dropout during training, where 50\% of the training scans use occupancy instead of reflectivity. During evaluation, reflectivity is used if the acquisition sensor matches the training sensor; otherwise, occupancy is applied.

Results for this method were computed with SRU-Net model and compared with and without reflectivity across all datasets, as shown in~\autoref{tab:dropout_reflec}. We observe a significant improvement in the source-to-source case and comparable results in most domain generalization scenarios relative to the model trained without reflectivity.

\begin{table}[h!]
    \centering
    \resizebox{1.0\linewidth}{!}{
    \begin{tabular}{l|c|ccccccc}        
    \textbf{SRU-Net Model} & SK & SK32 & P64 & PFF  & SP & W & NS & PL3D  \\
    \hline
    With reflectivity &\textbf{61,9}& 57,0 & \textbf{44,7} & 16,5& 45,2 & 26,0 &39,6  & 30,7  \\
    Without reflectivity & 58,6 & 54,0 & 44,2 & \textbf{22,2}& 45,3& \textbf{33,1}& 42,7 & 33,1  \\
    With dropout &61,4 & \textbf{57,4}  & 43,3 & 17,2& \textbf{46,4}& 32,2& \textbf{45,4} &\textbf{33,9}    \\
    \end{tabular}
    }
    \vspace{1mm}
    \caption{Comparison of domain generalization performance for SRU-Net with three different reflectivity strategies: with reflectivity, without reflectivity, and using the dropout strategy. All models were trained on SemanticKITTI.}
    \label{tab:dropout_reflec}
\end{table}

Overall, this is an interesting strategy, but it does not consistently outperform the approach without reflectivity. For simplicity, we will train our model without reflectivity for the remainder of this work.

With all implementation choices and planned experiments outlined, we can now discuss the domain generalization results.

\subsection{LiDAR domain generalization from SemanticKITTI}

We present the domain generalization results for various models and our 3DLabelProp approach, organized into two parts:~\autoref{tab:DG_from_SK} provides mIoU comparisons across several domain shifts, and~\autoref{tab:SSDGPL3DSK} offers mIoU and per-class IoU comparisons on ParisLuco3D.

\begin{table}[h!]
    \centering
    \resizebox{1.0\linewidth}{!}{
    \begin{tabular}{lc|c|cccccc}        
    \textbf{Model} & Input type & SK & SK32 & P64 & PFF  & SP & W & NS \\
    \hline
    CENet~\cite{cenet} & Range & 58,8 & 39,1 & 13,3 & 4,9& 27,9 & 7,4& 5,0  \\
    Helix4D~\cite{helix4D} & 4D sequence & 60,0 & 53,2 & 27,7 & 14,2& 36,0& N/A& 34,3  \\
    KPConv~\cite{kpconv}& Point & 58,3 & 52,7 & 32,7 & \cellcolor{yellow!30}21,1& 39,1 & 17,5 & \cellcolor{red!30}46,7 \\
    SRU-Net~\cite{mink}& Voxel & 58,6 & \cellcolor{yellow!30}54,0 & \cellcolor{orange!30}44,2 & \cellcolor{orange!30}22,2 & \cellcolor{yellow!30}45,3 & \cellcolor{orange!30}33,1 & 42,7 \\
    SPVCNN~\cite{spvnas}& Voxel \& point & \cellcolor{red!30}62,3 & \cellcolor{orange!30}57,4 & \cellcolor{yellow!30}40,2 & 19,4 & \cellcolor{orange!30}45,4 & \cellcolor{yellow!30}29,7& \cellcolor{yellow!30}45,1  \\
    C3D~\cite{cylinder3d}&Cylind. voxel& \cellcolor{yellow!30}60,7 & 53,1 & 18,4 & 6,5& 41,9& 18,9& 32,7 \\
    3DLabelProp & Pseudo-dense& \cellcolor{orange!30}61,9 & \cellcolor{red!30}61,7  & \cellcolor{red!30}57,3  & \cellcolor{red!30}59,3 & \cellcolor{red!30}47,2 & \cellcolor{red!30}39,4 & \cellcolor{orange!30}45,6 \\
    \end{tabular}
    }
    \vspace{1mm}
    \caption{Domain generalization performances (mIoU) of LSS models trained on SemanticKITTI and evaluated on six target datasets.}
    \label{tab:DG_from_SK}
\end{table}

\begin{table*}[h!]
    \centering
    \resizebox{0.9\linewidth}{!}{
    \begin{tabular}{l|ccccccccccccccccc|c}
        \multicolumn{1}{c}{\textbf{Model}} & \mcrot{1}{c}{Car} & \mcrot{1}{c}{Bicycle} & \mcrot{1}{c}{Motorcycle}& \mcrot{1}{c}{Truck}&\mcrot{1}{c}{Other-vehicle} &\mcrot{1}{c}{Person} & \mcrot{1}{c}{Road}& \mcrot{1}{c}{Parking}&\mcrot{1}{c}{sidewalk} & \mcrot{1}{c}{Other-ground}&\mcrot{1}{c}{building} & \mcrot{1}{c}{Fence}&\mcrot{1}{c}{Vegetation} &\mcrot{1}{c}{Trunk} & \mcrot{1}{c}{Terrain}& \mcrot{1}{c}{Pole}& \mcrot{1}{c}{Traffic-sign} & \textbf{mIoU} \\ 
        \hline 
        CENet~\cite{cenet} & 2,2 & 0 & 0 & 0,4 & 0 & 0 & 1,6 & 0 & 6,8 & 0 & 10,1 & 13,1 & 10,6 & 0 & 1,1 & 2,6 & 1,3 & 2,9 \\
        Helix4D~\cite{helix4D} & 18,2 & 0,2 & 2,1 & 0,3 & 4,7 & 3,0 & 55,1 & \cellcolor{orange!30}2,7 & 42,7 & \cellcolor{red!30}2,1 & 40,5 & \cellcolor{orange!30}22,1 & 51,0 & 27,3 & 7,7 & 27,9 & \cellcolor{orange!30}35,9 & 20,2\\
        KPConv~\cite{kpconv} & 39,8 & \cellcolor{yellow!30}7,4 & 9,1 & 0,3 & 5,1 & \cellcolor{red!30}30,6 & 8,1 & 0,1 & 41,5 & 0,7 & 58,5 & 11,7 & 66,9 & \cellcolor{orange!30}49,3 & \cellcolor{yellow!30}14,0 & 25,6 & 6,8 & 22,1\\
        SRU-Net~\cite{mink} & \cellcolor{red!30}69,5 & \cellcolor{orange!30}9,3 & \cellcolor{orange!30}15,4 & \cellcolor{red!30}4,1 & \cellcolor{red!30}32,1 & 20,5 & \cellcolor{red!30}71,4 & \cellcolor{yellow!30}1,0 & \cellcolor{orange!30}66,7 & \cellcolor{yellow!30}1,5 & \cellcolor{orange!30}67,3 & \cellcolor{yellow!30}18,1 & \cellcolor{orange!30}71,4 & \cellcolor{yellow!30}44,2 & \cellcolor{orange!30}15,7 & \cellcolor{red!30}40,2 & 19,0 & \cellcolor{orange!30}33,3 \\
        SPVCNN~\cite{spvnas} & \cellcolor{orange!30}66,7 & 7,0 & \cellcolor{yellow!30}14,0 & \cellcolor{orange!30}4,0 & \cellcolor{yellow!30}18,9 & \cellcolor{yellow!30}21,8 & \cellcolor{yellow!30}66,6 & 0,2 & \cellcolor{red!30}67,0 & 0,1 & \cellcolor{yellow!30}66,3 & 13,0 & \cellcolor{red!30}71,6 & 43,2 & 10,8 & \cellcolor{yellow!30}38,3 & \cellcolor{yellow!30}25,4 & \cellcolor{yellow!30}31,5 \\
        Cylinder3D~\cite{cylinder3d} & 46,4 & 4,6 & 5,8 & 0,3 & 15,2 & 11,3 & 58,9 & \cellcolor{red!30}3,9 & 57,2 & \cellcolor{orange!30}1,7 & 65,8 & \cellcolor{red!30}36,6 & 54,5 & 24,4 & 10,8 & 31,7 & 3,8 & 25,5\\
        3DLabelProp & \cellcolor{yellow!30}64,5 & \cellcolor{red!30}14,1 & \cellcolor{red!30}25,7 & \cellcolor{yellow!30}3,1 & \cellcolor{orange!30}27,1 & \cellcolor{orange!30}29,6 & \cellcolor{orange!30}70,3 & \cellcolor{orange!30}2,7 & \cellcolor{yellow!30}57,9 & 0,2 & \cellcolor{red!30}72,1 & 17,9 & \cellcolor{yellow!30}70,6 & \cellcolor{red!30}50,5 & \cellcolor{red!30}27,5 & \cellcolor{orange!30}38,8 & \cellcolor{red!30}37,1 & \cellcolor{red!30}35,9 \\
    \end{tabular}
    }
    \vspace{1mm}
    \caption{Domain generalization performances on ParisLuco3D dataset with LSS models trained on SemanticKITTI.}
    \label{tab:SSDGPL3DSK}
\end{table*}

The main conclusion from these tables is the sensitivity of all native models to domain shifts, particularly to substantial sensor shifts. In contrast, 3DLabelProp demonstrates strong resilience to sensor shifts and shows consistent improvements in domain generalization, showing the effectiveness of pseudo-dense point clouds.

With this observation in mind, we can delve deeper into the quantitative results to understand how the different models respond to various domain shifts. First, voxel-based methods tend to show resilience to scene and appearance shifts (e.g., SK $\rightarrow$ P64, SK $\rightarrow$ SP). Cylinder3D deviates from this trend due to the PointNet embedded in its architecture, which causes it to overfit the training domain and significantly reduces its domain generalization performance. Under sensor shift, all traditional methods struggle, especially with strong sensor shifts (as seen in SK $\rightarrow$ P64 vs. SK $\rightarrow$ PFF), where none manage to extract meaningful information. Range-based methods are particularly sensitive to sensor shift (e.g., SK $\rightarrow$ SK32). Point-based methods also struggle with sensor shift but exhibit strong resilience to appearance shifts (e.g., SK $\rightarrow$ NS). In class-wise results, these models perform exceptionally well in recognizing pedestrians across all datasets.

As anticipated from the analysis in~\autoref{tab:accumu}, 3DLabelProp demonstrates strong resilience to sensor shifts and remains effective under other types of domain shifts. Unlike naive pseudo-dense approaches, 3DLabelProp also achieves satisfactory source-to-source performance, with only a -0.4\% reduction compared to the best method. In cases of simple sensor shifts (SK $\rightarrow$ SK32), 3DLabelProp achieves a notably higher mIoU than the second-best method (+4.3\%) and maintains the smallest performance gap to source-to-source results (-0.2\%). For significant sensor shifts, it is the only method able to extract meaningful information from PandaFF (+37.1\% compared to the second-best method) with the smallest gap to Panda64. For appearance shifts, 3DLabelProp performs significantly better on Panda64 (+13.1\% compared to the second-best method) and Waymo (+6.3\% compared to the second-best method) and moderately better on SemanticPOSS (+1.8\%). The single exception is the SK $\rightarrow$ NS scenario (-1.1\% compared to the best method), where KPConv slightly outperforms 3DLabelProp. Overall, 3DLabelProp proves to be a highly competitive method for semantic segmentation and domain generalization.

Previous analyses provided a macroscopic view, helping us understand each method's general domain generalization tendencies. Using~\autoref{tab:SSDGPL3DSK}, which evaluates models on ParisLuco3D (where label annotations match those of SemanticKITTI), we can examine each method’s performance at a class level. Voxel-based methods excel in recognizing the ground and vehicles, while, KPConv is particularly effective at detecting pedestrians. Other methods show less convincing results. 3DLabelProp performs especially well in identifying bikes, pedestrians, and structures.

\begin{figure*}[h!]
        \centering
        \subfloat[Ground Truth]{
            \includegraphics[width=.23\linewidth]{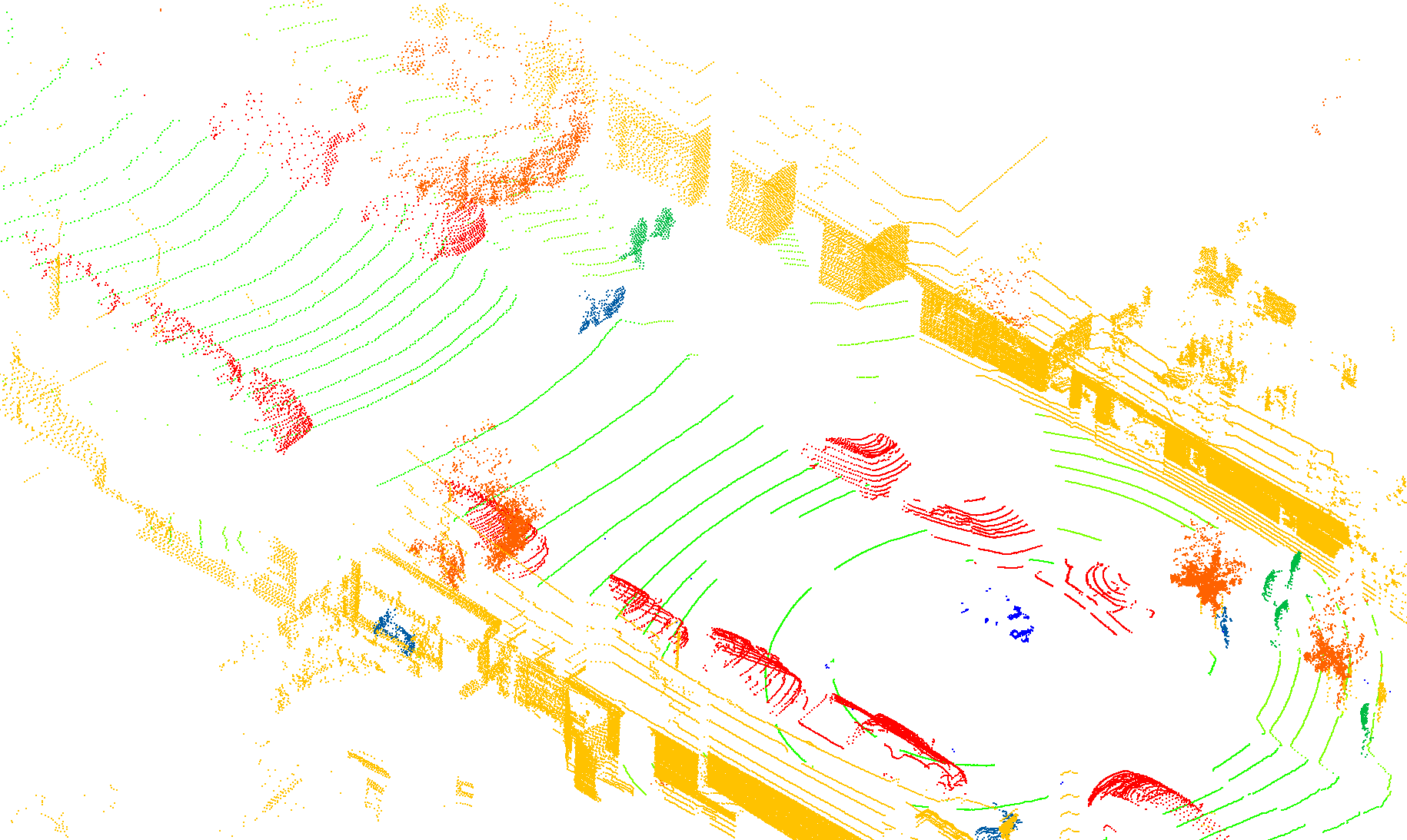}
        }
        \subfloat[KPConv]{
            \includegraphics[width=.23\linewidth]{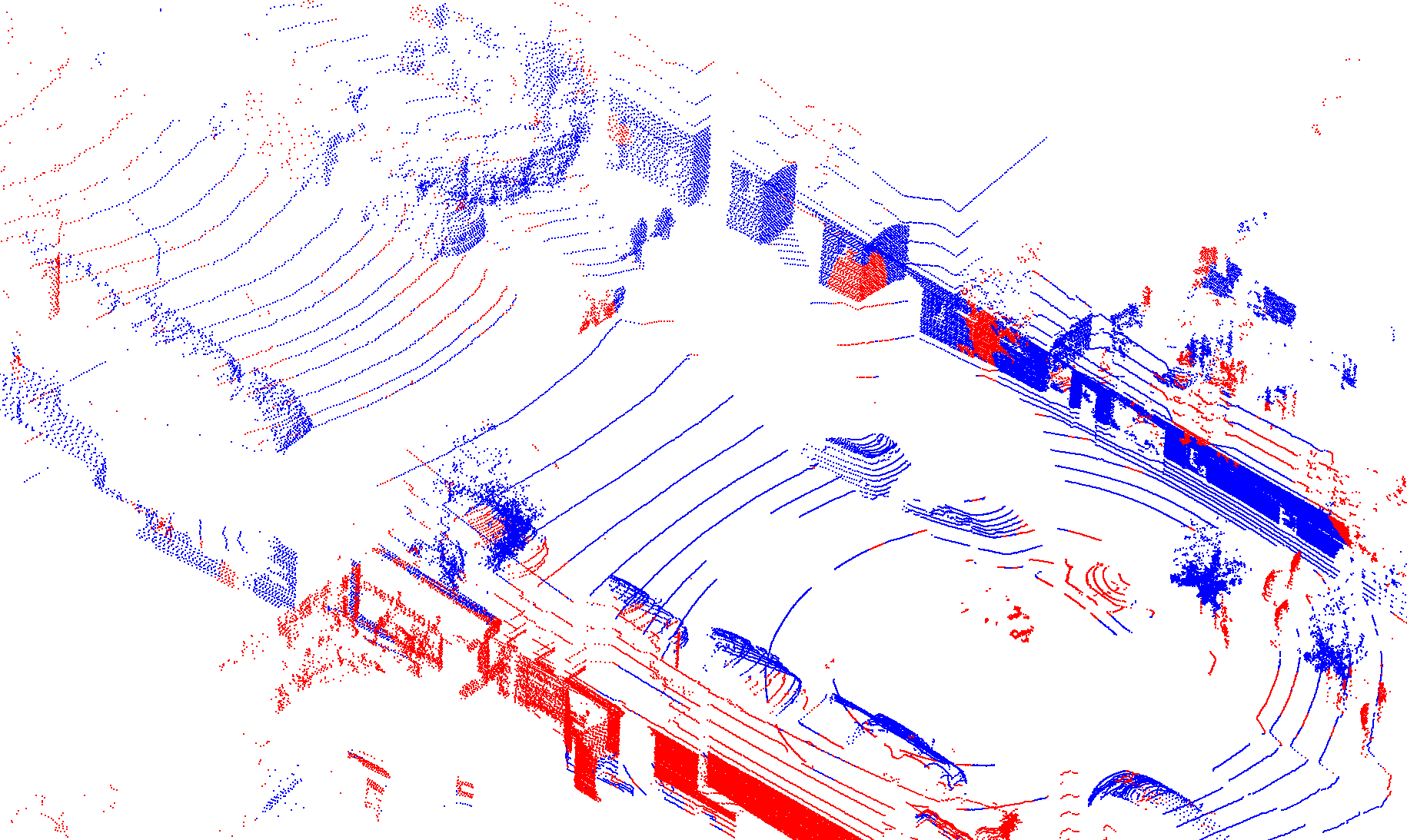}
        } 
        \subfloat[SPVCNN]{
            \includegraphics[width=.23\linewidth]{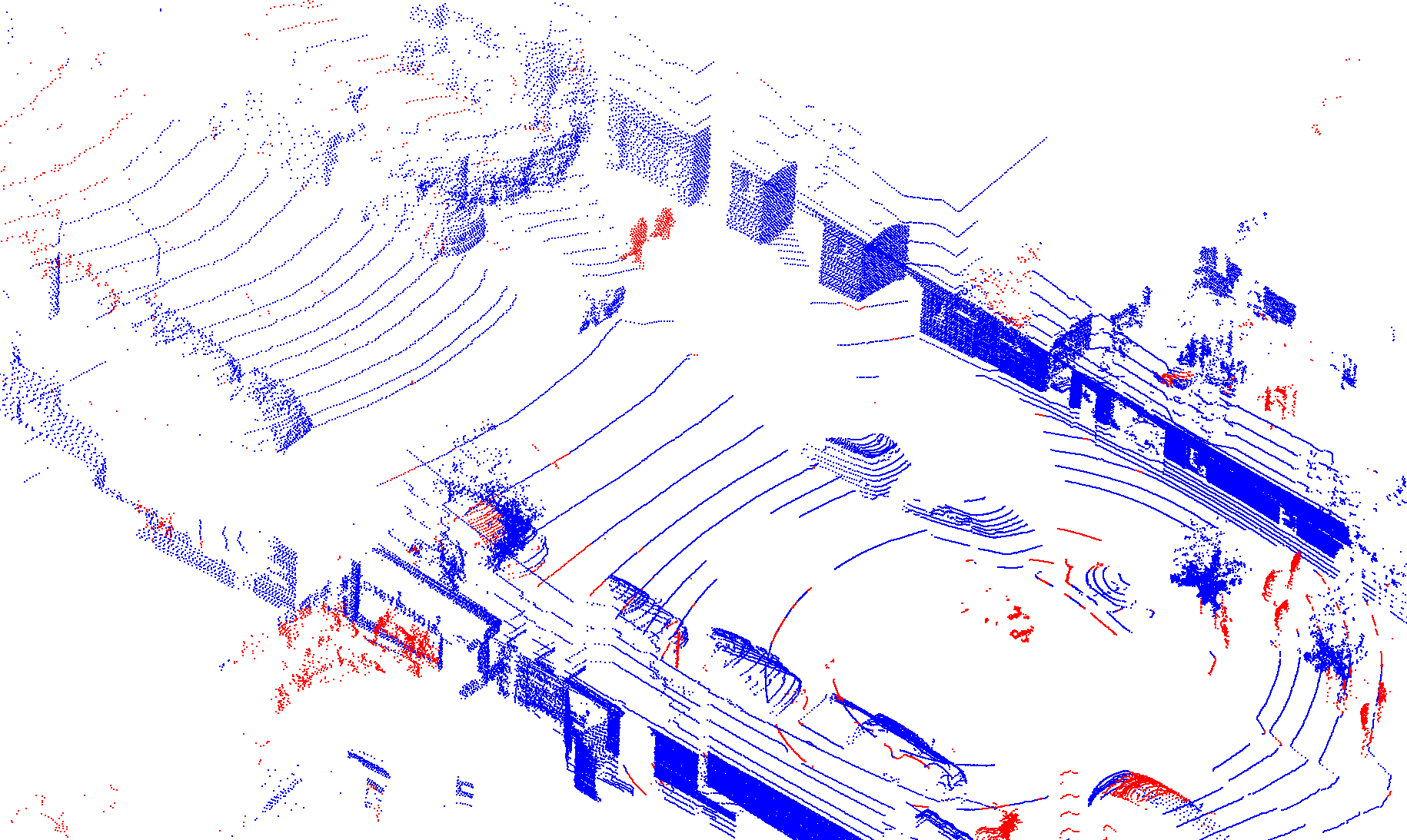}
        }
        \subfloat[3DLabelProp (Ours)]{
            \includegraphics[width=.23\linewidth]{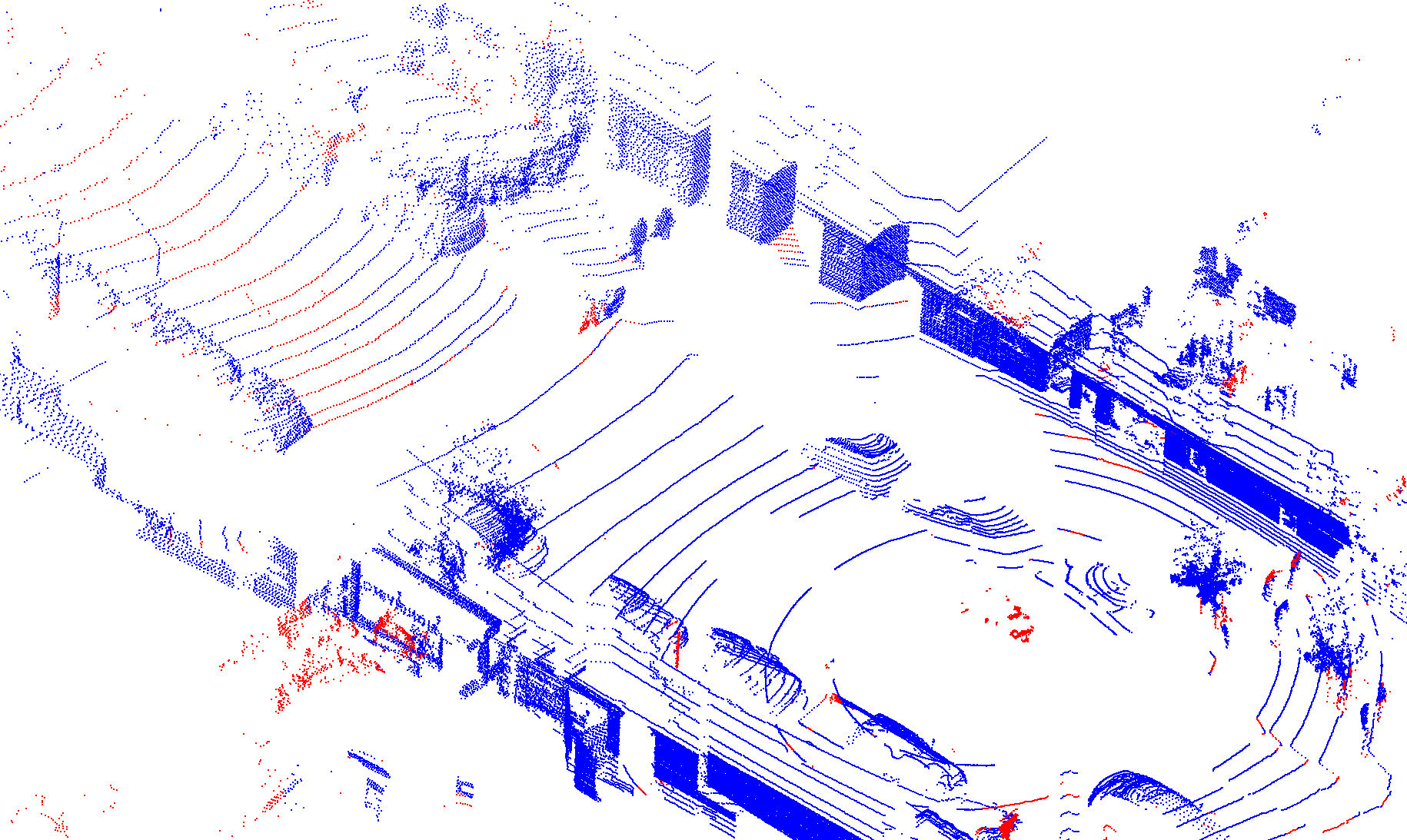}
        } 
        
        \subfloat[Ground Truth]{
            \includegraphics[width=.23\linewidth]{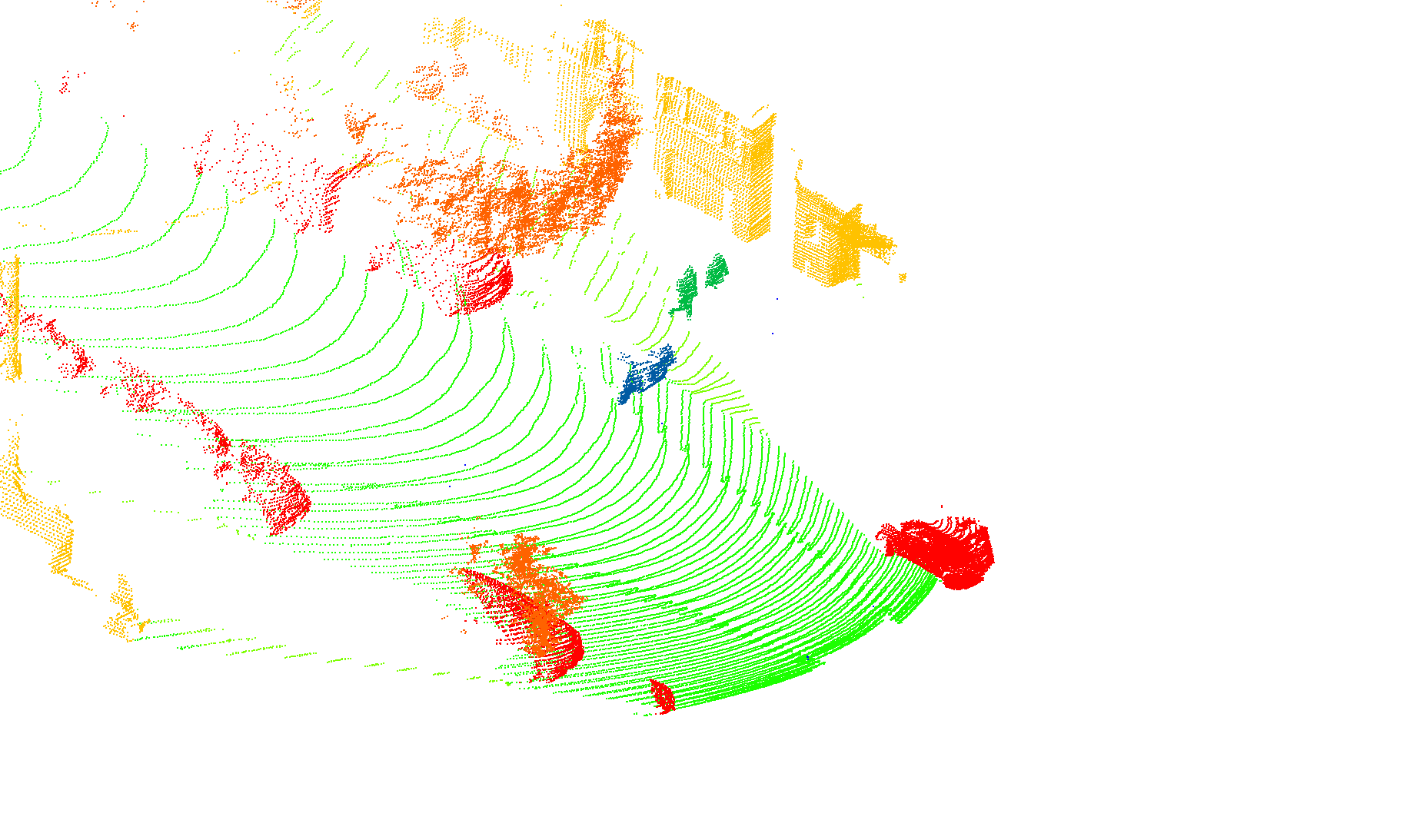}
        }
        \subfloat[KPConv]{
            \includegraphics[width=.23\linewidth]{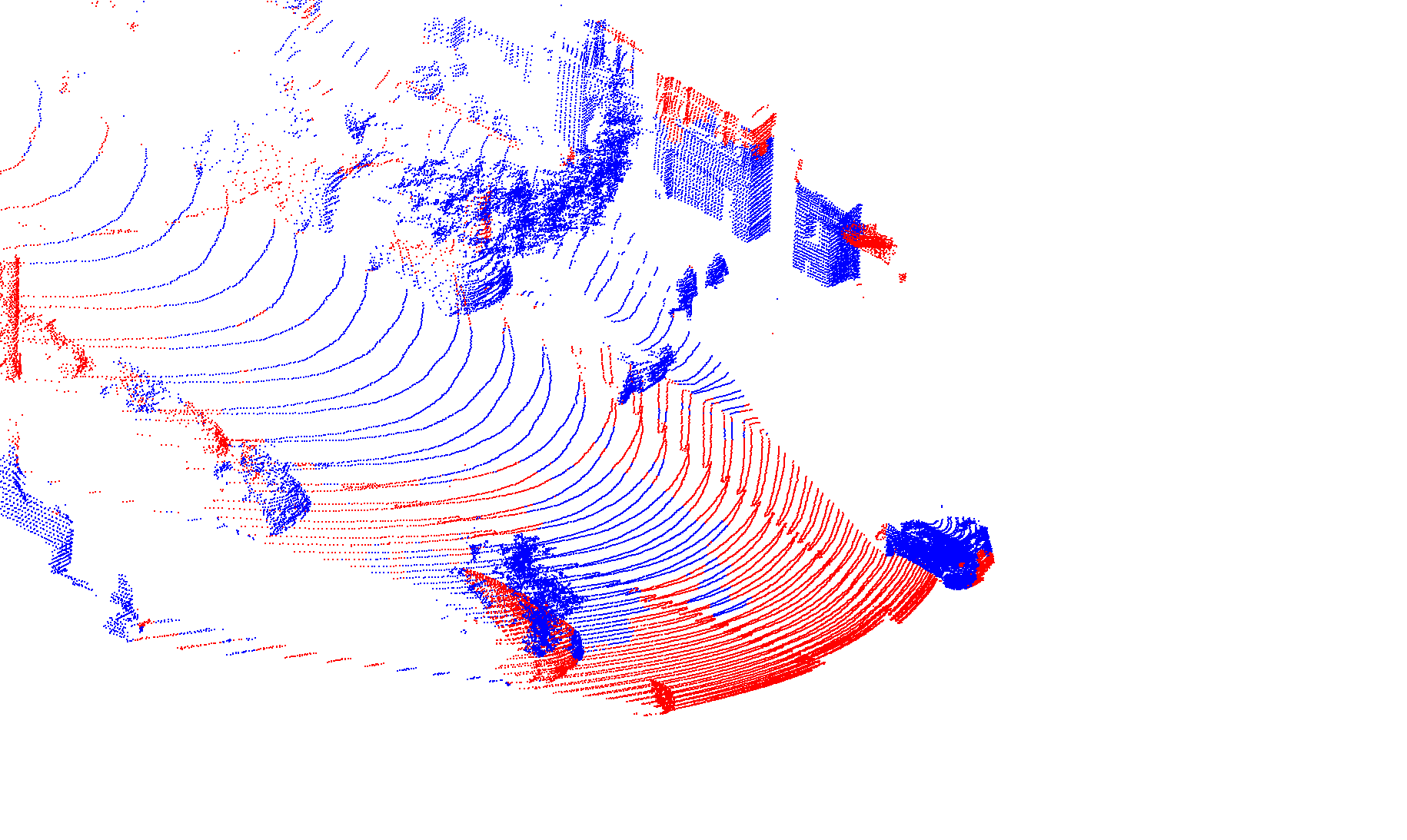}
        } 
        \subfloat[SPVCNN]{
            \includegraphics[width=.23\linewidth]{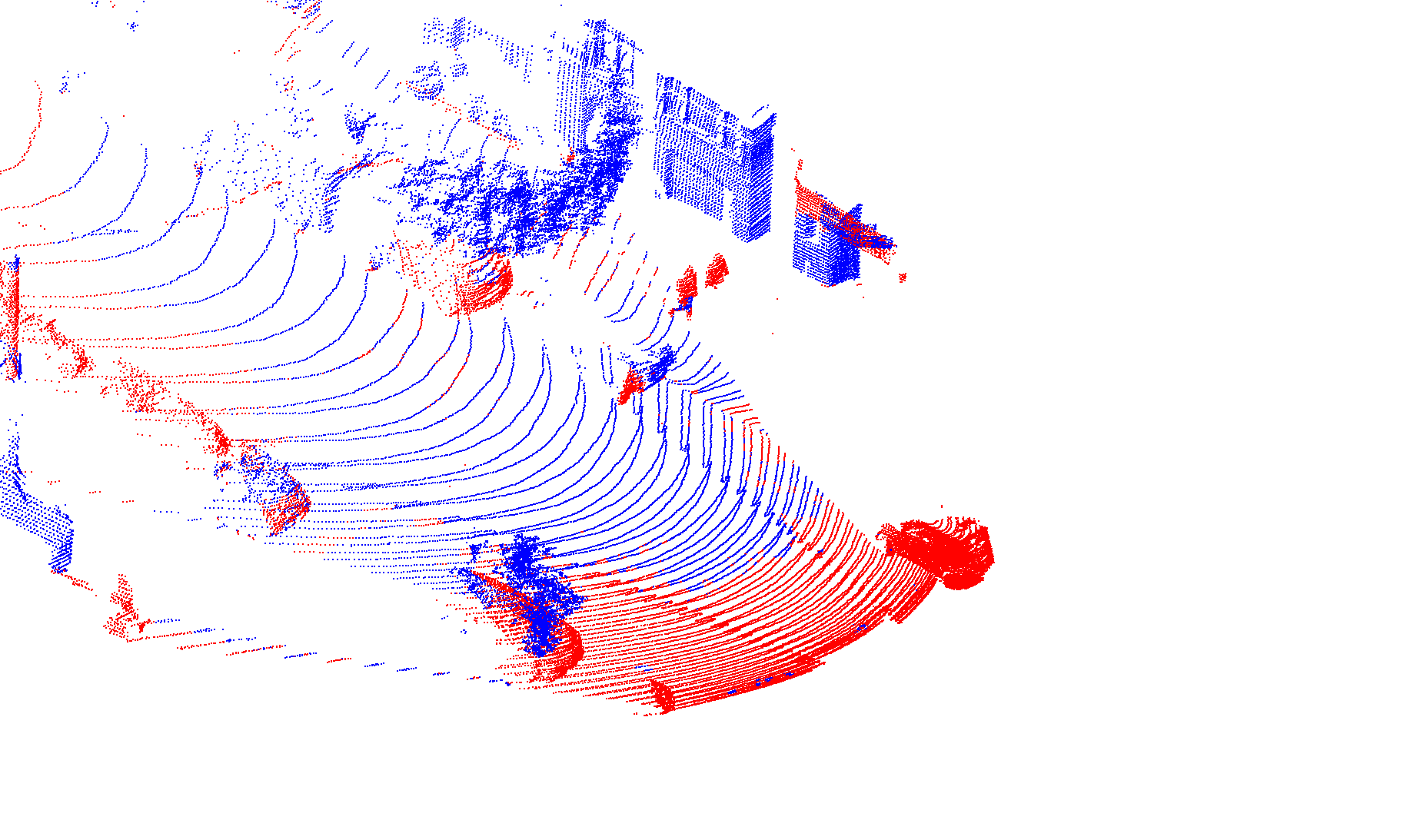}
        }
        \subfloat[3DLabelProp (Ours)]{
            \includegraphics[width=.23\linewidth]{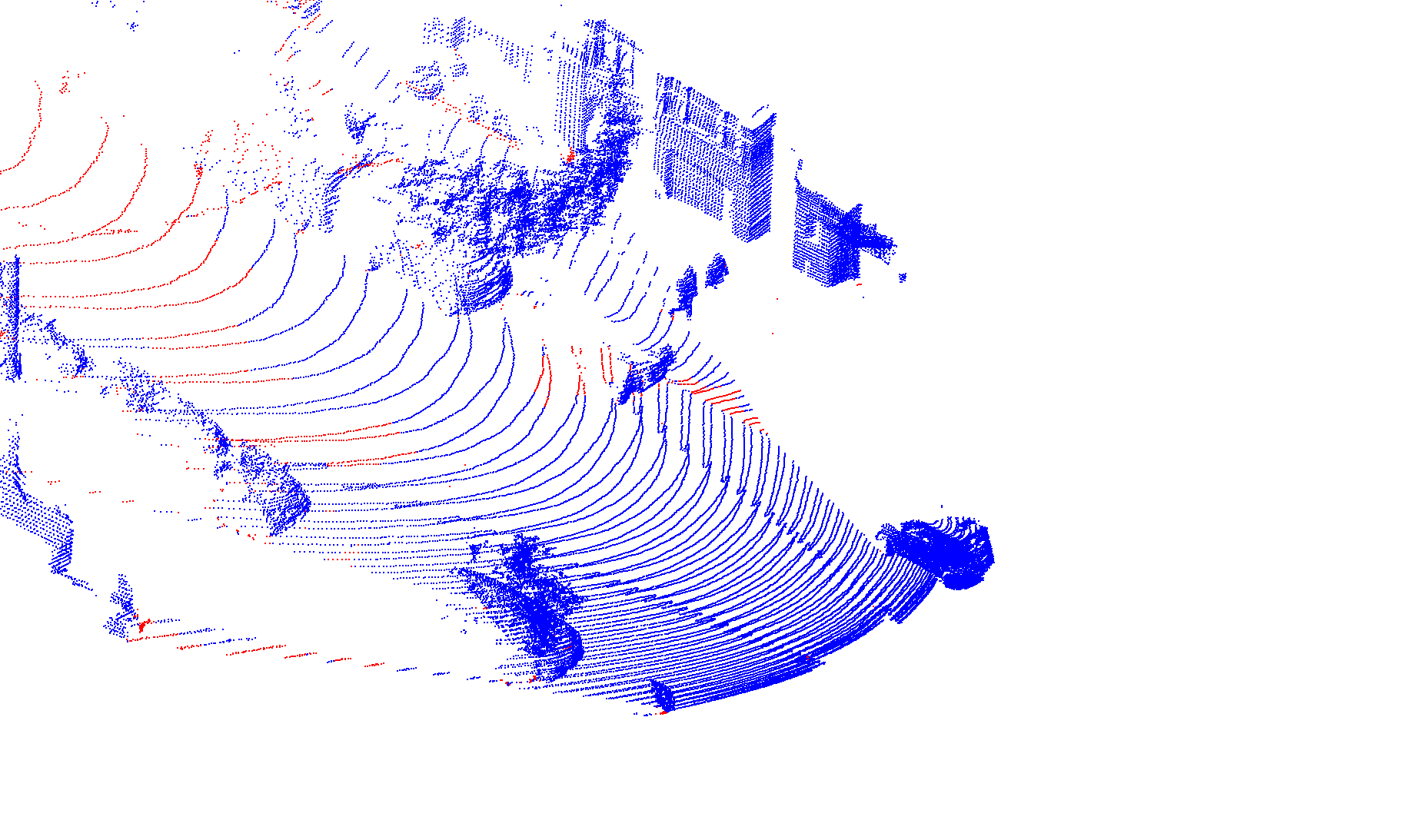}
        } 
        \caption{Qualitative results for KPConv~\cite{kpconv}, SPVCNN~\cite{spvnas}, and 3DLabelProp trained on SemanticKITTI and tested on Panda64 (top row) and PandaFF (bottom row), two different LiDAR sensors from the same PandaSet dataset. Correctly segmented points are shown in blue, while errors are shown in red. }
\label{fig:quali1}
\end{figure*}

We present qualitative semantic segmentation results in~\autoref{fig:quali1} for the models KPConv, SPVCNN, and 3DLabelProp, trained on SemanticKITTI and tested on Panda64 and PandaFF, two different LiDAR sensors from the same PandaSet dataset. Blue points indicate correct semantics, while red points represent errors. We can see that KPConv and SPVCNN methods exhibit significant segmentation errors when evaluated on the PandaFF dataset, which uses a solid-state LiDAR sensor very different from the Velodyne HDL64 LiDAR sensor used in the training dataset. These errors occur even on simple classes like the road, particularly near the sensor, an issue that could be critical for safety in autonomous driving applications. 3DLabelProp is the only method capable of accurately segmenting points close to the sensor for PandaFF.

\subsection{LiDAR domain generalization from nuScenes}

When using nuScenes as the training source, we observe similar conclusions to those from the SemanticKITTI case, as the result patterns are very comparable. Our analysis is again divided into two parts: \autoref{tab:DG_from_NS} presents mIoU comparisons across various domain shifts, while~\autoref{tab:SSDGPL3DNS} provides per-class IoU comparisons on ParisLuco3D.

\begin{table}[h!]
\centering
\resizebox{1.0\linewidth}{!}{
\begin{tabular}{lc|c|cccccc}
    \textbf{Model} & Input type  & NS & SK & SK32 & P64 & PFF  & SP & W \\
    \hline
    CENet~\cite{cenet}& Range & 69,1 & 10,0 & 49,6 & 9,8 & 6,0 & 3,3 & 3,5\\
    Helix4D~\cite{helix4D} & 4D sequence & \cellcolor{yellow!30}69,3 & 40,0 &44,1   & 13,6 & 7,6  & 45,2 & N/A \\
    KPConv~\cite{kpconv} & Point & 63,1 & 44,9 & 50,6 & 25,0 & \cellcolor{orange!30}16,9 & 60,7 & 15,2\\
    SRU-Net~\cite{mink} & Voxel & 66,3 & \cellcolor{yellow!30}46,3 & \cellcolor{yellow!30}52,4& \cellcolor{yellow!30}33,1 & 9,8 & \cellcolor{yellow!30}61,5 & \cellcolor{yellow!30}23,6 \\
    SPVCNN~\cite{spvnas} & Voxel \& point & 67,2 & \cellcolor{orange!30}49,4 & \cellcolor{orange!30}53,2 & \cellcolor{orange!30}43,7 & \cellcolor{yellow!30}11,1 & \cellcolor{red!30}64,8 & \cellcolor{orange!30}37,2 \\
    C3D~\cite{cylinder3d} & Cylind. voxel& \cellcolor{orange!30}70,2 & 31,7 & 46,1 & 15,8 & 4,7 & 42,8 & 12,7 \\
    3DLabelProp & Pseudo-dense & \cellcolor{red!30}71,5 & \cellcolor{red!30}59,8 & \cellcolor{red!30}62,1 & \cellcolor{red!30}66,2 & \cellcolor{red!30}70,0  & \cellcolor{orange!30}64,6 & \cellcolor{red!30}47,1 \\
\end{tabular}
}
\vspace{1mm}
\caption{Domain generalization performances (mIoU) of LSS models trained on nuScenes and evaluated on six target datasets.}
\label{tab:DG_from_NS}
\end{table}

In~\autoref{tab:DG_from_NS}, 3DLabelProp consistently proves to be the best generalization method and even achieves superior source-to-source segmentation results. Leveraging pseudo-dense point clouds is particularly advantageous when training on lower-resolution input as nuScenes.

\begin{table*}[h!]
    \centering
    \resizebox{0.9\linewidth}{!}{
    \begin{tabular}{l|cccccccccccccccc|c}
        \multicolumn{1}{c}{\textbf{Model}} & \mcrot{1}{c}{barrier}& \mcrot{1}{c}{bicycle}& \mcrot{1}{c}{bus}& \mcrot{1}{c}{car}& \mcrot{1}{c}{construction-vehicle}& \mcrot{1}{c}{motorcycle}& \mcrot{1}{c}{pedestrian}& \mcrot{1}{c}{traffic-cone}& \mcrot{1}{c}{trailer}& \mcrot{1}{c}{truck}& \mcrot{1}{c}{driveable-surface}& \mcrot{1}{c}{other-flat}& \mcrot{1}{c}{sidewalk}&\mcrot{1}{c}{terrain} & \mcrot{1}{c}{manmade}& \mcrot{1}{c}{vegetation}& \textbf{mIoU} \\ 
        \hline 
        CENet~\cite{cenet} & \cellcolor{yellow!30}4,1 & \cellcolor{orange!30}4,7 & 35,7 & 61,6 & 1,6 & \cellcolor{yellow!30}22,7 & 51,6 & 0 & 0 & 6,1 & \cellcolor{orange!30}77,4 & \cellcolor{red!30}22,5 & \cellcolor{orange!30}56,7 & \cellcolor{yellow!30}13,3 & 81,1 & 66,6 & 31,6 \\
        Helix4D~\cite{helix4D} & 1,0 & 0,4 & 9,2 & 29,2 & 0,1 & 2,4 & 9,3 & 0 & 0 & 0,1 & 57,4 & \cellcolor{yellow!30}5,8 & 40,8 & 13,0 & 71,5 & 65,3 & 19,2 \\
        KPConv~\cite{kpconv} & \cellcolor{orange!30}4,2 & 0,1 & 2,1 & 8,3 & 0,2 & 1,9 & 11,4 & 0 & 0 & 1,7 & 8,3 & 0 & 1,7 & 3,7 & 73,2 & 65,7 & 11,4  \\
        SRU-Net~\cite{mink} & 1,8 & 0,4 & \cellcolor{yellow!30}48,4 & \cellcolor{orange!30}78,1 & \cellcolor{yellow!30}3,9 & 10,6 & \cellcolor{yellow!30}51,7 & 0,3 & 0 & \cellcolor{red!30}20,4 & \cellcolor{yellow!30}72,7 & 2,5 & \cellcolor{yellow!30}47,4 & \cellcolor{orange!30}14,3 & \cellcolor{yellow!30}83,5 & \cellcolor{orange!30}83,2 & \cellcolor{orange!30}32,5 \\
        SPVCNN~\cite{spvnas} & 1,7 & \cellcolor{yellow!30}0,8 & \cellcolor{orange!30}49,5 & \cellcolor{yellow!30}66,0 & \cellcolor{orange!30}5,4 & \cellcolor{orange!30}27,1 & \cellcolor{orange!30}55,9 & \cellcolor{yellow!30}0,3 & 0 & \cellcolor{orange!30}15,4 & 69,3 & 1,4 & 42,8 & 11,8 & \cellcolor{orange!30}84,8 & 82,1 & \cellcolor{yellow!30}32,1  \\
        Cylinder3D~\cite{cylinder3d} & \cellcolor{yellow!30}0,3 & 0 & 5,0 & 31,5 & 0,4 & 0,1 & 17,4 & \cellcolor{red!30}1,2 & 0 & \cellcolor{yellow!30}14,0 & 13,3 & 0,1 & 25,5 & 5,8 & 77,3 & \cellcolor{yellow!30}82,4 & 17,1 \\
        3DLabelProp & \cellcolor{red!30}9,6 & \cellcolor{red!30}24,1 & \cellcolor{red!30}50,0 & \cellcolor{red!30}80,6 & \cellcolor{red!30}14,3 & \cellcolor{red!30}65,2 & \cellcolor{red!30}78,1 & \cellcolor{orange!30}0,7 & 0 & 7,8 & \cellcolor{red!30}79,2 & \cellcolor{orange!30}6,8 & \cellcolor{red!30}68,3 & \cellcolor{red!30}29,2 & \cellcolor{red!30}93,2 & \cellcolor{red!30}89,4 & \cellcolor{red!30}43,5 \\
    \end{tabular}
    }
    \vspace{1mm}
    \caption{Domain generalization performances on ParisLuco3D dataset with LSS models trained on nuScenes.}
    \label{tab:SSDGPL3DNS}
\end{table*}

The analysis of ParisLuco3D in~\autoref{tab:SSDGPL3DNS} differs slightly from that of SemanticKITTI. Unlike SemanticKITTI, nuScenes is captured in urban area and has greater scene similarity to ParisLuco3D. Additionally, there is no sensor shift between nuScenes and ParisLuco3D, as they use the exact same LiDAR sensor. ParisLuco3D also includes annotations that enable evaluation of models trained on nuScenes with exactly the same classes. In~\autoref{tab:SSDGPL3DNS}, range-based methods show decent generalization, especially in recognizing ground types. 3DLabelProp achieves even better performance than in the SemanticKITTI $\rightarrow$ ParisLuco3D case, demonstrating its adaptability to various types of domain shifts and not only sensor shift. As before, it excels particularly in recognizing bikes and pedestrians.

In conclusion, semantic segmentation methods used naively for domain generalization tend to yield underwhelming results. However, voxel-based methods show more promise compared to others, while range-based methods are effective only when there is no sensor shift. 3DLabelProp demonstrates strong capabilities, especially when trained on lower-resolution datasets. In the next section, 3DLabelProp will be compared with other domain generalization methods.

\section{Comparison with domain generalization methods}

\subsection{Comparison with C\&L}

The previous sections highlighted the effectiveness of using pseudo-dense point clouds for domain generalization, showing consistent improvement over traditional semantic segmentation methods. However, the prior benchmark did not include comparisons with other 3D domain generalization methods. In this section, we address this by providing such a comparison.

One important point to note is that comparing with other generalization methods is challenging, as each defines its own evaluation label set, making direct comparison difficult.

First, we compare our method with Complete \& Label~\cite{completelabel}, as it is fundamentally the closest to ours. As a reminder, their approach involves extracting the canonical domain using a completion model. In~\autoref{tab:complete}, we present the unsupervised domain adaptation results from C\&L (access to target data without labels), given that their domain generalization analysis is quite limited (considering only two classes: vehicles and pedestrians).

\begin{table}[h!]
    \centering
    \begin{tabular}{l|ccc|ccc}
     \multicolumn{1}{c}{}&\multicolumn{3}{c}{SK $\rightarrow$ NS} & \multicolumn{3}{c}{NS $\rightarrow$ SK} \\
     Method & SK & NS & \% drop & NS & SK &  \% drop \\
     \hline 
        C\&L~\cite{completelabel}& 50,2 & 31,6 & \textbf{-37\%} & 54,4 & 33,7 & -38\%\\
        3DLabelProp & \textbf{69,0} & \textbf{42,7} & -38\% & \textbf{66,5} & \textbf{50.5} & \textbf{-24\%} \\
    \end{tabular}
    \vspace{1mm}
    \caption{Comparison of 3DLabelProp with C\&L~\cite{completelabel} on the C\&L label set (mIoU), trained on SemanticKITTI and evaluated on nuScenes, and vice versa. C\&L results are their unsupervised domain adaptation results.}
    \label{tab:complete}
\end{table}

Although a direct comparison of domain generalization results is not entirely fair due to the differing backbones, we observe that the relative performance drop (compared to source-to-source segmentation) is smaller for our method. This demonstrates its effectiveness and supports our claim that geometry-based canonical domain recovery is more robust than learning-based approaches.

\subsection{Comparison with LiDOG and DGLSS}

Next, we compare our approach with more recent and competitive domain generalization methods: LIDOG~\cite{lidog} and DGLSS~\cite{domgen}.

\begin{table}[h!]
    \centering
    \begin{tabular}{l|ccc|ccc}
        \multicolumn{1}{c}{}&\multicolumn{3}{c}{SK $\rightarrow$ NS}&\multicolumn{3}{c}{NS $\rightarrow$ SK} \\ 
        Method & SK & NS & \% drop & NS & SK &  \% drop \\
        \hline 
        LIDOG~\cite{lidog} & 61,5 & 34,9& -43\% &48,5 &41,2  & -15\%\\
        3DLabelProp & \textbf{83,0}& \textbf{58,8} & \textbf{-29\%} & \textbf{82,4} & \textbf{73,9} & \textbf{-10\%}  \\
    \end{tabular}
    \vspace{1mm}
    \caption{Comparison of 3DLabelProp with LiDOG approach on the LiDOG label set (mIoU), trained on SemanticKITTI and evaluated on nuScenes, and vice versa.}
    \label{tab:lidog}
\end{table}

The quantitative comparison with LiDOG is provided in~\autoref{tab:lidog}. Like C\&L, LiDOG uses a shallower deep architecture, leading to lower source-to-source performance while employing a simplified label set that boosts 3DLabelProp’s performance. Therefore, we include the relative decrease to allow a fair comparison between the methods. 3DLabelProp consistently outperforms LiDOG.

DGLSS~\cite{domgen} is a domain generalization method for LiDAR semantic segmentation that operates on an larger label set than previous domain generalization methods. It employs the same SRU-Net architecture as in our previous analysis (\autoref{tab:DG_from_SK}), making it a competitive approach.

\begin{table}[h!]
    \centering
    \begin{tabular}{l|c|cc}
        Method & SK & NS & W  \\ 
        \hline
        DGLSS \cite{domgen}& 59,6& \textbf{44,8} & 40,7 \\
        3DLabelProp & \textbf{74,7} & 44,2 & \textbf{43,6} \\
    \end{tabular}
    \vspace{1mm}
    \caption{Comparison of 3DLabelProp with DGLSS approach on the DGLSS label set (mIoU). All methods are trained with SemanticKITTI.}
    \label{tab:dglss}
\end{table}

The quantitative comparison with DGLSS is shown in~\autoref{tab:dglss}. We achieve comparable domain generalization results (-0.6\% on nuScenes and +2.9\% on Waymo) while obtaining significantly higher source-to-source performance. The notable improvement in results for 3DLabelProp for SemanticKITTI, compared to~\autoref{tab:DG_from_SK}, is due to the chosen label set, which excludes bicyclists and motorcyclists, the two most challenging classes. In contrast, DGLSS shows a decline in source-to-source performance compared to SRU-Net from~\autoref{tab:DG_from_SK}. It should be noted that the datasets on which DGLSS was evaluated (nuScenes and Waymo) are ones where our approach performs comparably to the SRU-Net method (see~\autoref{tab:DG_from_SK}), as they include multiple domain shifts.

In conclusion, 3DLabelProp is a highly competitive method compared to other domain generalization approaches. Our previous conclusions hold: 3DLabelProp achieves strong domain generalization results without compromising source-to-source performance.

\section{Ablation study of 3DLabelProp}

\subsection{Influence of geometric parameters}

In presenting 3DLabelProp, we introduced several geometric hyperparameters. This section examines these parameters to demonstrate 3DLabelProp's robustness to hyperparameter settings and to provide guidance on selecting them.

We identified three key parameters: $d_p$, the distance propagation for labels; $K_c$, the number of clusters during K-means clustering; and $N_s$, the number of past scans used to create the pseudo-dense point cloud. For all previously shown results, regardless of the training and evaluation sets, we used: $d_p = 0.30m$, $K_c = 20$, and $N_s = 20$.

To assess the impact of each parameter, we conducted a one-at-a-time analysis using nuScenes as the training dataset and SemanticKITTI as the target dataset, varying each parameter individually while keeping the others constant. The results are shown in~\autoref{tab:ab_geomtrique}.

%We only studied 3 parameters as $v_s$ follows the industry standard and $V_c$ is a memory parameter. The higher $V_c$ is, the better the results are but the higher the memory consumption is. All trainings were done on nuScenes.

\begin{table}[h!]
        \centering
        \resizebox{1.0\linewidth}{!}{
        \begin{tabular}{ccc|c|cc}
            $d_p$ (m) & $K_c$ & $N_s$ & mIoU$_{\mathcal{L}_{NS}}^{NS}$ & mIoU$_{\mathcal{L}_{NS\cap SK}}^{SK}$ & Inference speed (Hz)\\
            \noalign{\vskip 0.5mm}
            \hline 
                \cellcolor{black!10}0,10 & \cellcolor{black!10}- & \cellcolor{black!10}- & \cellcolor{black!10}72,4 & \cellcolor{black!10}60,2 & \cellcolor{black!10}0,6\\ 
                0,30 & 20 & 20 & 71,5 & 59,8 & 1,2\\
                \cellcolor{black!10}0,60 & \cellcolor{black!10}- & \cellcolor{black!10}- & \cellcolor{black!10}69,1 & \cellcolor{black!10}57,2 & \cellcolor{black!10}1,5\\ 
                \hline
                \cellcolor{black!10}- & \cellcolor{black!10}5 & \cellcolor{black!10}- & \cellcolor{black!10}71,3 & \cellcolor{black!10}61,5 & \cellcolor{black!10}1,3\\ 
                0,30 & 20 & 20 & 71,5 & 59,8 & 1,2\\
                \cellcolor{black!10}- & \cellcolor{black!10}40 & \cellcolor{black!10}- & \cellcolor{black!10}68,8 & \cellcolor{black!10}59,7 & \cellcolor{black!10}0,9\\
                \hline
                \cellcolor{black!10}- & \cellcolor{black!10}- & \cellcolor{black!10}5 & \cellcolor{black!10}67,4 & \cellcolor{black!10}60,0 & \cellcolor{black!10}1,5\\
                \cellcolor{black!10}- & \cellcolor{black!10}- & \cellcolor{black!10}10 & \cellcolor{black!10}70,9 & \cellcolor{black!10}60,2 & \cellcolor{black!10}1,5\\
                0,30 & 20 & 20 & 71,5 & 59,8 & 1,2\\
                \cellcolor{black!10}- & \cellcolor{black!10}- & \cellcolor{black!10}40 & \cellcolor{black!10}67,9 & \cellcolor{black!10}59,5 & \cellcolor{black!10}1,0\\ 
        \end{tabular}
        }
        \vspace{1mm}
        \caption{Impact of the geometric parameters of 3DLabelProp trained on nuScenes (NS) and tested on SemanticKITTI (SK). In gray are the standard parameters of the method.}
        \label{tab:ab_geomtrique}
    \end{table}

The first observation is that even with suboptimal parameters ($d_p=0.60m$ or $N_s=5$), 3DLabelProp achieves satisfactory performance for both source-to-source and domain generalization, demonstrating the method's robustness to geometric parameter settings.

The most affected metric is processing speed, with a threefold difference between the fastest and slowest parameter sets. We selected $d_p=0.30m$ as a balance between speed and performance. Smaller values improve results by confining propagation to closer neighbors, but since the propagation step primarily drives the speed-up, reducing it significantly impacts processing speed.

$K_c$ follows a similar pattern: smaller values result in larger clusters, offering extensive contextual information. However, this increases point cloud size and memory usage, which then requires careful monitoring. A setting of $K_c=5$ caused memory issues with SemanticKITTI, leading us to choose $K_c=20$ by default.

Thus far, context has been the primary factor driving performance improvements. Accordingly, we might expect that increasing $N_s$ would further enhance results. However, in practice, this is not the case due to the 'trail effect' previously discussed. When the time window becomes too large, an excess of trails introduces noisy neighbors to newly sampled points within them. A balance must be struck between providing stable contextual information and limiting trail formation, leading us to select $N_s=20$.

\subsection{Influence of the backbone}

Since 3DLabelProp utilizes KPConv, it operates at a slower pace. It is therefore worthwhile to explore whether a different deep learning backbone could improve both speed and performance. In~\autoref{tab:backbone}, we evaluate SRU-Net as an alternative backbone. SRU-Net was selected for its high-quality pseudo-dense results and the efficiency of its single-scan implementation.

\begin{table}[h!]
    \centering
     \resizebox{1.0\linewidth}{!}{
    \begin{tabular}{l|c|ccccccc}
        Backbone & SK & SK32 & P64 & PFF  & SP & W & NS  & PL3D  \\
        \hline   
        KPConv~\cite{kpconv}  & \textbf{61,9}&\textbf{61,7}  &\textbf{57,3}  & \textbf{59,3}& \textbf{47,2}& \textbf{39,4}& \textbf{45,6}& \textbf{35.9} \\ 
        SRU-Net~\cite{mink} & 53.3&48.6& 51.3 &56.9& 46.1& 35.1& 40.0 & 33.9 \\ 
        \end{tabular}
        }
        \vspace{1mm}
        \caption{Domain generalization results depending on the backbone of 3DLabelProp, either KPConv (by default) or SRU-Net. All models are trained on SemanticKITTI.}
    \label{tab:backbone}
\end{table}

Overall, the results are underwhelming. While they are satisfactory for domain generalization compared to naive approaches, they fall significantly short of those achieved with the KPConv backbone. Feeding small point clusters into SRU-Net negatively impacts its performance. Voxel-based methods typically enable long-range interactions due to their network depth, which is not feasible with smaller point clouds. 3DLabelProp was specifically designed for use with KPConv, making it challenging to adapt effectively to other deep learning approaches.

\section{Limitations}

Despite several acceleration strategies, 3DLabelProp remains below real-time requirements (10 Hz for SemanticKITTI and 20 Hz for nuScenes). In~\autoref{tab:vitesse_exec_3DLabelProp}, we compare 3DLabelProp with KPConv on single scans and KPConv on pseudo-dense point clouds. 3DLabelProp consistently outperforms KPConv pseudo-dense and achieves similar inference speeds to KPConv, though it still falls short of real-time processing, which we leave for future research. Unlike KPConv and SRU-Net on pseudo-dense clouds, which encounter memory issues in certain cases (see~\autoref{tab:accumu}), 3DLabelProp avoids memory constraints thanks to a clustering step that creates smaller point clouds for processing. Overall, 3DLabelProp is a competitive pseudo-dense method in terms of both memory and speed.

\begin{table}[h!]
    \centering
    \begin{tabular}{l|cc}
        \textbf{Method} & Speed on SK (Hz)  & Speed on NS (Hz) \\ 
        \hline
        KPConv~\cite{kpconv} & 0,6 & 1,3\\
        KPConv pseudo-dense~\cite{kpconv} & 0,1 & 0,3\\ 
        3DLabelProp & 0,2 & 1,2\\
    \end{tabular}
    \caption{Inference speed comparisons between KPConv on single scan, KPConv on pseudo-dense point clouds and 3DLabelProp.}
    \label{tab:vitesse_exec_3DLabelProp}
\end{table}

\section{Conclusion}

In this work, we proposed a benchmark for state-of-the-art LiDAR semantic segmentation methods to evaluate their effectiveness in domain generalization, aiming to clarify the strengths and limitations of different approaches. Alongside the benchmark, we provide a comprehensive methodology for studying domain generalization by formalizing various domain shifts and analyzing them across a wide range of datasets.

Additionally, we conducted an in-depth analysis of the benefits and limitations of using pseudo-dense point clouds (a representation that benefit from performance and robustness of state-of-the-art LiDAR odometries) for semantic segmentation, demonstrating their promising results in domain generalization.

Lastly, we introduced a new 3D semantic segmentation domain generalization method, called 3DLabelProp, which utilizes pseudo-dense point clouds to minimize sensor discrepancies across acquisition devices.
3DLabelProp leverages geometry to propagate static labels in high-confidence areas while using deep networks to predict labels on point clusters in dynamic regions. This approach demonstrates resilience not only to highly different LiDAR sensors (e.g., from the Velodyne HDL64 in SemanticKITTI to the solid-state PandarGT in PandaSet) but also robustness to other domain shifts. However, the approach remains too slow for real-time applications (primarily due to the KPConv network) and could benefit from faster neural networks for dense point clouds in future research work.
%This method also shows competitive performance for moving object segmentation.

{\small
\bibliographystyle{IEEEtran}
\bibliography{bib.bib}
}

\newpage

\vspace{11pt}

\begin{IEEEbiographynophoto}{Jules Sanchez}
received his PhD degree in computer science and robotics at Mines Paris - PSL University in 2023. His research topics mainly focus on LiDAR semantic segmentation and domain generalization for autonomous vehicles, with publications in international conferences in Computer Vision and Robotics.
\end{IEEEbiographynophoto}

\begin{IEEEbiographynophoto}{Jean-Emmanuel Deschaud}
received his PhD degree in computer science and robotics in 2010.
He is currently an Associate Professor at the Centre for Robotics at Mines Paris - PSL University. His research focuses on LiDAR SLAM, neural networks for point cloud processing, and 3D perception for autonomous vehicles, with publications in international conferences and journals in the fields of Computer Vision and Robotics.
\end{IEEEbiographynophoto}

\begin{IEEEbiographynophoto}{François Goulette}
graduated from Mines Paris with an engineering degree in 1992 and a PhD in computer science and robotics in 1997. He obtained a Habilitation to Conduct Research from Sorbonne University in 2009. He worked for a few years as a research engineer at the French electrical company EDF and then as an assistant, associate, and full Professor at Mines Paris - PSL University. Since 2022, he has served as a full professor and deputy director of the "U2IS" Lab at ENSTA Paris - Institut Polytechnique de Paris. His research interests are 3D perception, mobile mapping, photogrammetry, 3D data processing for robotics, autonomous vehicles, and other applications.
\end{IEEEbiographynophoto}

\vfill

\end{document}